\documentclass[11pt]{article}
\usepackage[T1]{fontenc}
\usepackage[utf8]{inputenc}
\usepackage{lmodern}
\usepackage{microtype}
\usepackage[margin=1in]{geometry}
\usepackage{setspace}
\usepackage{graphicx}
\usepackage{booktabs}
\usepackage{tabularx}
\usepackage{longtable}
\usepackage{array}
\usepackage{amsmath,amssymb}
\usepackage[numbers,sort&compress]{natbib}
\usepackage{caption}
\usepackage{xcolor}
\usepackage{xurl}
\usepackage{placeins}
\usepackage[hidelinks]{hyperref}
\newcolumntype{Y}{>{\raggedright\arraybackslash}X}
\newcolumntype{C}[1]{>{\centering\arraybackslash}p{#1}}

\title{\bfseries EndoRSI: Physical-Evidence-Driven Recursive Improvement of Endoscopic Robot Policies}
\author{%
Chi Kit Ng\textsuperscript{1,\textdagger}, Yidong Zhang\textsuperscript{1,\textdagger}, Tianchun Wu\textsuperscript{1},\\
Siu Hing Lui\textsuperscript{1}, Hanlin Zhao\textsuperscript{2}, Huxin Gao\textsuperscript{1} and Hongliang Ren\textsuperscript{1,*}\\[0.5em]
\small \textsuperscript{1}Department of Electronic Engineering, The Chinese University of Hong Kong,\\[-0.1em]
\small Hong Kong SAR, China\\[0.3em]
\small \textsuperscript{2}Department of Computer Science and Engineering, The Chinese University of Hong Kong,\\[-0.1em]
\small Hong Kong SAR, China\\[0.4em]
\small \textsuperscript{\textdagger}These authors contributed equally: Chi Kit Ng and Yidong Zhang.\\
\small \textsuperscript{*}Correspondence: Hongliang Ren (\href{mailto:hlren@ee.cuhk.edu.hk}{hlren@ee.cuhk.edu.hk}).
}
\date{}
\hypersetup{pdftitle={EndoRSI: Physical-Evidence-Driven Recursive Improvement of Endoscopic Robot Policies},pdfauthor={Chi Kit Ng et al.}}

\begin{document}
\maketitle
\singlespacing
\begin{abstract}
Artificial intelligence is beginning to automate scientific discovery, yet physical robot-policy research still relies on scientists and engineers to interpret experimental failures and decide how controllers should be revised. We present EndoRSI (Endoscopic Recursive Self-Improvement), a human-governed, self-improving robotic research system in which the embedded EndoForge coding agent iteratively develops and refines endoscopic robot policies using evidence from physical experiments. Physical execution remains isolated within a safety-constrained runtime under human authorisation. In motorised endoscopy, validation evidence first caused the agent to reject an initial action representation, while subsequent physical experiments falsified an offline-supported insertion strategy by revealing hazardous closed-loop behaviour absent from the source data. Failed physical rollouts were then incorporated into policy revision; insertion success subsequently increased from 2/21 in the first generation to 17/21 in the fourth. The frozen final policy retained competence across visual and hardware shifts and was the best-performing autonomous policy across six independent ex-vivo porcine specimens. These results show that physical robot interaction can serve as experimental evidence for agent-driven robot-policy research.
\end{abstract}

\section*{Introduction}

Artificial intelligence (AI) is beginning to automate scientific discovery. Coding agents can inspect repositories, modify code and execute software tasks~\citep{yang2024sweagent}. AI-scientist systems extend such capabilities across ideation, experimentation, analysis and scientific writing~\citep{lu2026aiscientist}. Self-driving laboratories and robot scientists have closed experimental loops in materials science, chemistry and biology, allowing machine-generated hypotheses to be tested against physical measurements~\citep{macleod2022selfdriving,burger2020robotchemist,king2004robotscientist,king2009automation}. Separately, program-search systems have demonstrated autonomous discovery in mathematics~\citep{romeraparedes2024math}. These systems change the unit of automation from a single prediction to an experimental cycle: observe evidence, generate a hypothesis, implement a test, interpret the result and decide what to test next.

Despite these advances, end-to-end automation of evidence interpretation, policy revision and physical retesting remains uncommon in robot-policy development. Language models can generate robot programs and grounded plans~\citep{liang2023code,ahn2022saycan} and design rewards or sim-to-real configurations~\citep{ma2024eureka,ma2024dreureka}. A complementary challenge is to make the evidential chain of physical policy research auditable while retaining explicit human authority over deployment, outcome labelling and emergency intervention. Generalist vision--language--action (VLA) policies broaden the skills and embodiments represented by a controller~\citep{brohan2023rt2,driess2023palme,kim2025openvla,oneill2024openx,black2024pi0}, and robot learning provides powerful mechanisms for imitation, temporal action generation and experience reuse~\citep{chi2023diffusion,zhao2023act,ross2011dagger,kober2013rlrobotics,kumar2020cql,kostrikov2022iql}. Yet scientists and engineers still commonly determine whether a poor result calls for a new representation, new supervision or a different policy class. Closing this loop is especially difficult because physical evidence is sequential, safety constrained and produced by the policy's own actions~\citep{brunke2022safety}.

\begin{figure}[!htbp]
\centering
\includegraphics[width=\linewidth,height=0.70\textheight,keepaspectratio,trim={39.10bp 9.02bp 79.90bp 0.00bp},clip]{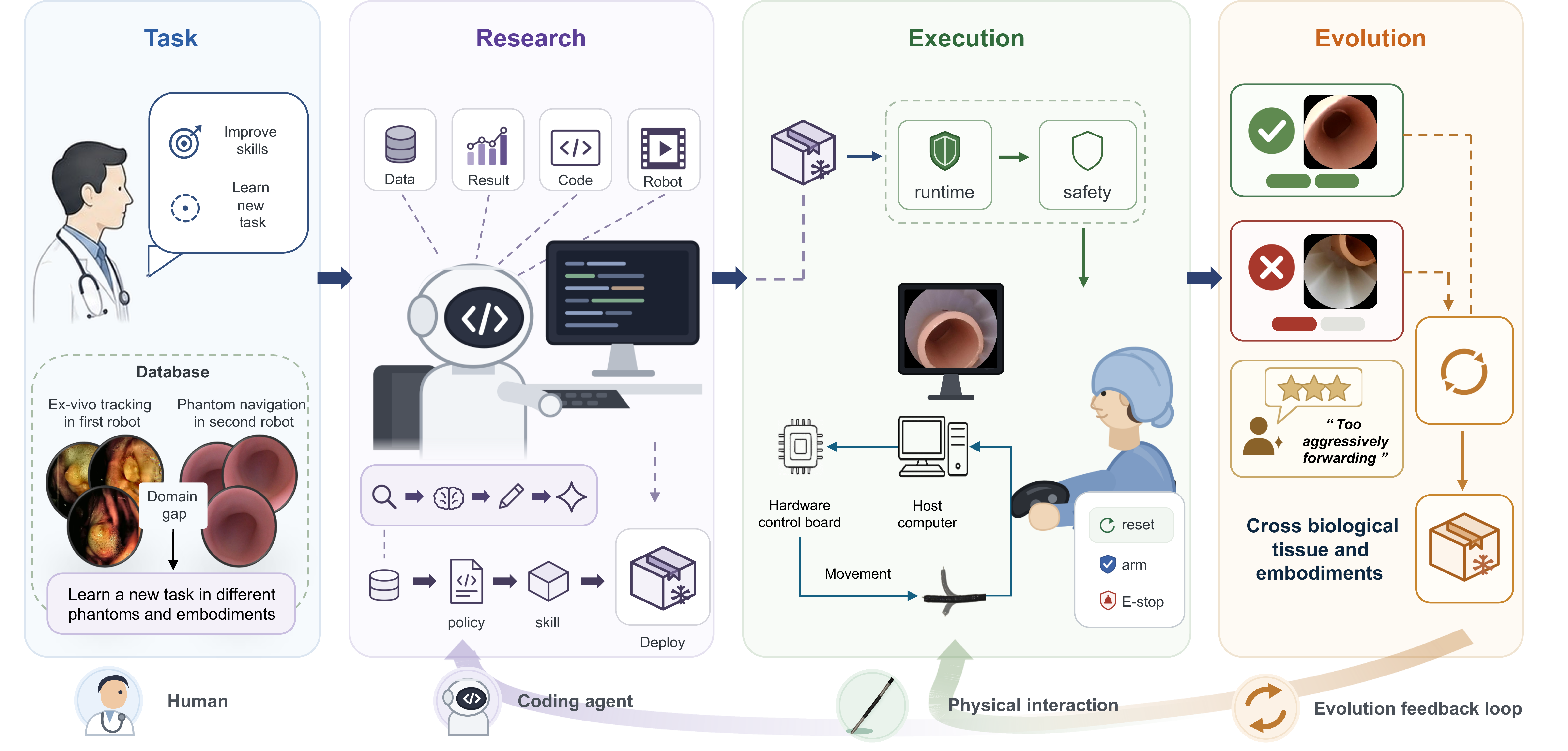}
\caption{\textbf{EndoForge connects task definition, policy research, physical execution and evidence-driven revision.} Existing tracking and navigation data provide complementary source skills. The coding agent analyses data, results and code and prepares candidate policies. Human-authorised execution occurs through a separate safety-constrained runtime. Physical outcomes and human feedback inform subsequent research and transfer evaluation.}
\label{fig:overview}
\end{figure}

EndoRSI (Endoscopic Recursive Self-Improvement) is a human-governed, self-improving robotic research system that enables its embedded EndoForge coding agent to iteratively develop and refine endoscopic robot policies using evidence from physical experiments. Within EndoRSI, EndoForge operates at the research timescale, analysing experimental outcomes, modifying policy implementations, constructing training supervision and proposing subsequent policy generations. Physical execution is performed by a separate safety-constrained robotic runtime under human authorisation, with experimental outcomes fed back to support further policy improvement. Here we show that a coding agent can use real-robot outcomes to reject and revise policy candidates within a human-bounded, safety-isolated research loop (Fig.~\ref{fig:overview}).

We tested this framework in motorised endoscopy, where confined geometry, contact and closed-loop state shifts make apparently reasonable offline policies particularly vulnerable to physical failure. Robotic autonomy has advanced endoscopy and minimally invasive surgery~\citep{yang2017medical,saeidi2022star,zhang2024copilot,martin2020magnetic}, while endoscopic AI and surgical-data systems provide increasingly rich representations and datasets~\citep{ali2022endoscopy,azagra2023endomapper,maierhein2017surgicaldatascience,openhembodiment2026}. These properties create a stringent test of the research loop: source skills can appear composable offline, yet contact and policy-induced state visitation can invalidate the composition within seconds.

We first tested whether validation evidence could redirect agent-designed policies on existing tracking and navigation skills. We then examined whether those skills could support a new insertion policy without insertion trajectories collected on the target task, and whether physical interaction could falsify a composition that appeared plausible offline. Finally, we tested whether failed physical rollouts could provide useful supervision for subsequent generations and whether the resulting policy transferred across deployment shifts and biological tissue.

\section*{Results}

\begin{figure}[!htbp]
\centering
\includegraphics[width=\linewidth,height=0.80\textheight,keepaspectratio,trim={0bp 0bp 0bp 0bp},clip]{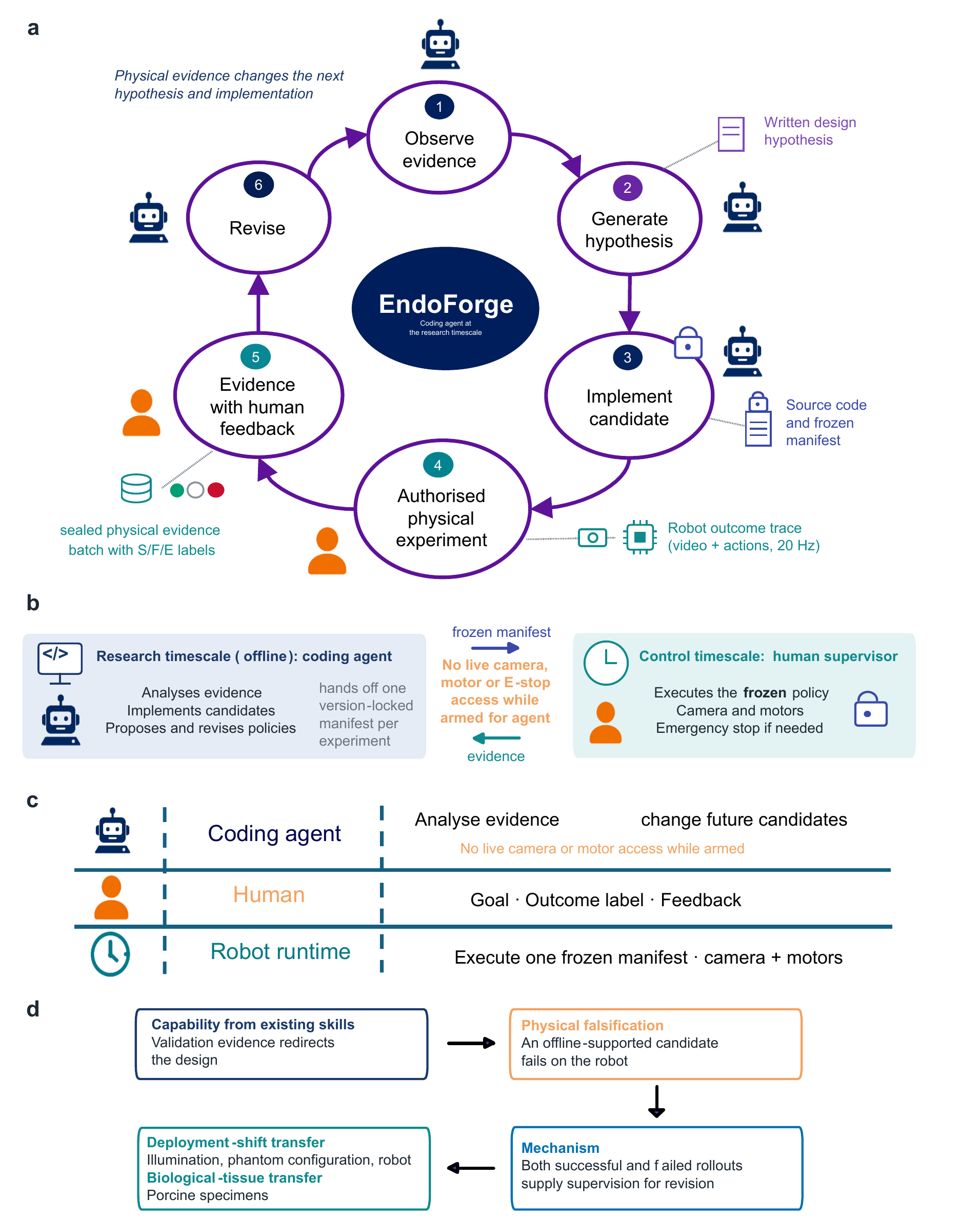}
\caption{\textbf{Physical evidence changes subsequent research decisions.} \textbf{a}, The research loop connects analysis, hypotheses, candidate implementation, authorised experiments, outcome feedback and revision. \textbf{b}, The coding agent acts offline; a separate runtime executes the fixed policy under human supervision. \textbf{c}, Division of responsibilities among the coding agent, human investigators and runtime. \textbf{d}, The experimental progression from source-skill validation to physical falsification, failure-derived supervision and transfer. Success, failure and emergency-stop outcomes are denoted S, F and E, respectively.}
\label{fig:researchloop}
\end{figure}

We evaluated whether physical robot experiments could provide actionable evidence for agent-driven policy development in motorised endoscopy. EndoForge operates at the research timescale: the coding agent analyses available data and experimental outcomes, implements candidate policies and proposes subsequent revisions, while a separate safety-constrained runtime executes only human-authorised policies (Figs.~\ref{fig:overview} and \ref{fig:researchloop}). We examined this research loop in four stages. First, we asked whether validation evidence could cause the agent to reject and redesign a policy trained from existing tracking and navigation data. Second, we tested whether physical interaction could falsify a new insertion strategy that appeared plausible from those source skills alone. Third, we investigated whether failed physical rollouts could provide useful supervision for subsequent policy generations. Finally, we held the evolved policy fixed and evaluated its transfer across visual, hardware and ex-vivo biological conditions.

\subsection*{Validation evidence redirects policy development}

EndoForge began with two complementary source skills collected on different robotic systems. Target tracking taught a medical endoscope to steer towards a visual target in an ex-vivo pig stomach, whereas navigation captured forward and bending behaviour from an industrial motorised endoscope traversing a colon phantom. Their combination was relevant to insertion because the new task required both alignment with an opening and controlled advancement through it. A third, smaller-diameter industrial endoscope was reserved for later hardware-transfer testing (Supplementary Fig.~1 and Supplementary Table~3). Each source dataset contained 100 episodes divided into 80 training, 10 validation and 10 held-out test episodes (Supplementary Table~5).

\begin{figure}[!htbp]
\centering
\includegraphics[width=\linewidth,height=0.70\textheight,keepaspectratio,trim={0.00bp 0.00bp 0.00bp 0.00bp},clip]{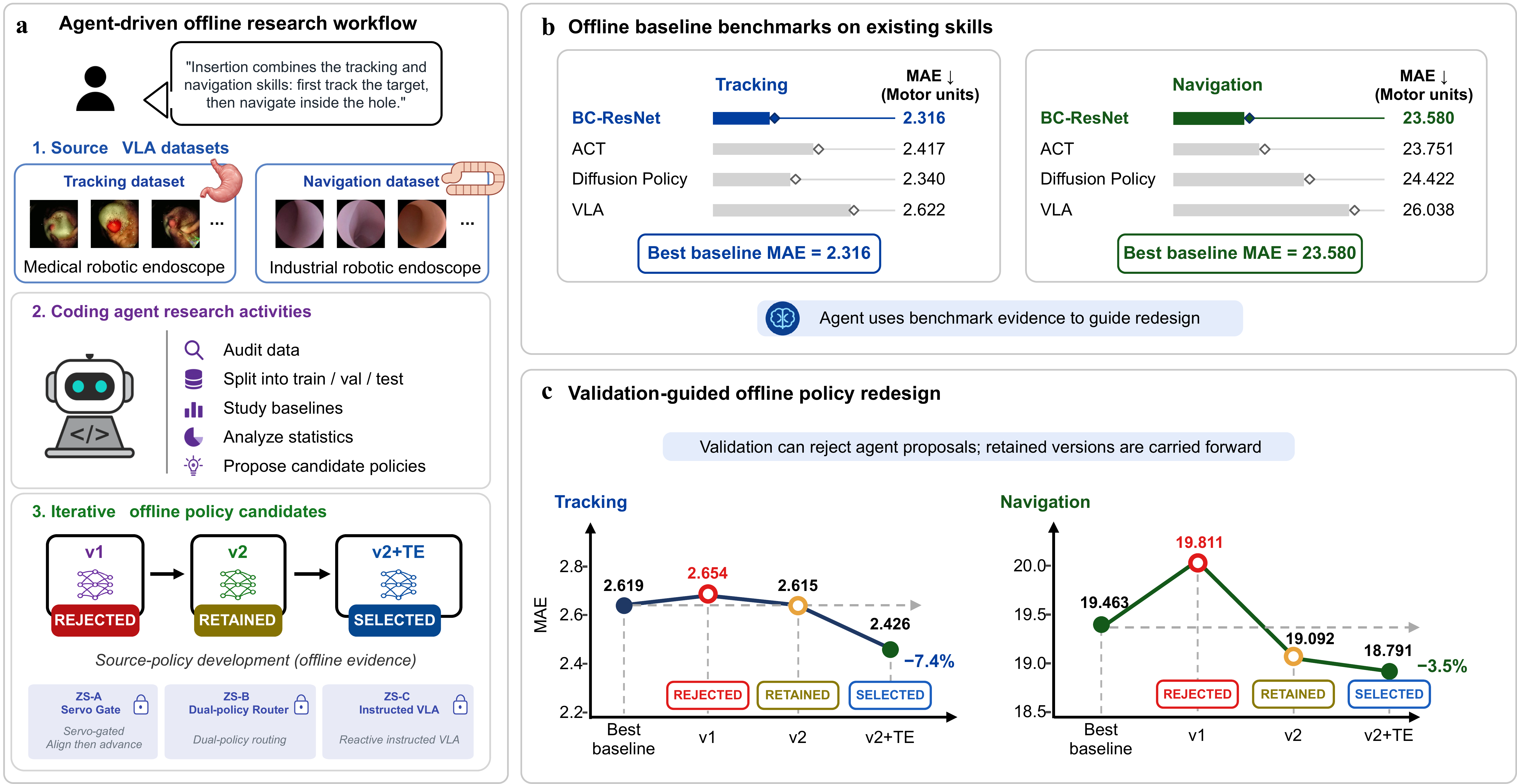}
\caption{\textbf{Validation evidence redirects policy design.} \textbf{a}, Source tracking and navigation tasks, coding-agent activities and candidate dispositions. \textbf{b}, Held-out image-conditioned reference comparisons. Diamond markers indicate the reported MAE values at the bar endpoints; thin horizontal lines lead to the numerical labels and do not indicate uncertainty intervals. \textbf{c}, Validation MAE during the progression from v1 to direct continuous prediction in version 2 (v2) and temporal ensembling (TE). Validation determined candidate selection; held-out evaluation quantified generalisation. The validation and held-out comparisons use different episodes and therefore different baseline values. Exact split-specific values and metrics are reported in Supplementary Note 3 and Supplementary Tables~7 and 8. MAE is expressed in motor units. ZS-B denotes the dual-policy router and ZS-C the instructed VLA policy.}
\label{fig:offline}
\end{figure}

To determine whether validation evidence could redirect policy design, the agent reproduced image-conditioned references and implemented a first candidate, version 1 (v1) (Fig.~\ref{fig:offline}). The v1 candidate stacked three frames and represented each future action as a discrete bin plus a continuous residual. On the validation split, relative to a convolutional neural network (CNN) behaviour-cloning reference, this representation increased tracking mean absolute error (MAE) from 2.619 to 2.654 motor units and navigation MAE from 19.463 to 19.811. The agent attributed the failure to a mismatch between argmax bin decoding and the continuous error objective, rejected v1 and replaced it with direct continuous prediction.

The revised version 2 (v2) policy used the current image and a temporal-difference image to regress continuous eight-step action chunks. Temporal ensembling (TE) then causally combined overlapping predictions. On held-out episodes, v2+TE reduced tracking MAE from 2.316 to 2.112 and navigation MAE from 23.580 to 21.683. The paired episode-level differences were $-0.204$ motor units (95\% confidence interval (CI), $-0.387$ to $-0.0004$) and $-1.916$ ($-3.377$ to $-0.322$), respectively, and all three independent training runs remained below the CNN reference on both tasks (Supplementary Table~8). Thus, validation evidence changed both the action representation and the decoding strategy before physical deployment.

An intentionally teacher-forced copy-previous diagnostic, which received the preceding ground-truth action at every step and ignored the image, achieved still lower error but could not be deployed autonomously. This result exposed strong temporal autocorrelation in the demonstrations: a deceptively low offline error could be obtained without visual control. Offline validation could therefore reject an inferior representation, but it could not establish competence in the closed-loop states created by the physical robot.

\subsection*{Physical interaction falsifies policy composition}

A policy composition that appeared justified by the source datasets failed catastrophically on the physical robot. Investigators proposed that insertion could be decomposed into tracking-based alignment followed by navigation-based advancement. The agent tested this hypothesis against the two source datasets. Target offset carried a usable directional bending signal, and forward motion was concentrated in low-bending navigation segments; by contrast, a seemingly intuitive relation between bending and the dark-lumen centroid was unsupported and was rejected. The complete coefficients, quartile statistics and candidate dispositions are reported in Supplementary Note~4 and Supplementary Table~9.

\begin{figure}[!htbp]
\centering
\includegraphics[width=\linewidth,height=0.80\textheight,keepaspectratio,trim={0bp 0bp 0bp 0bp},clip]{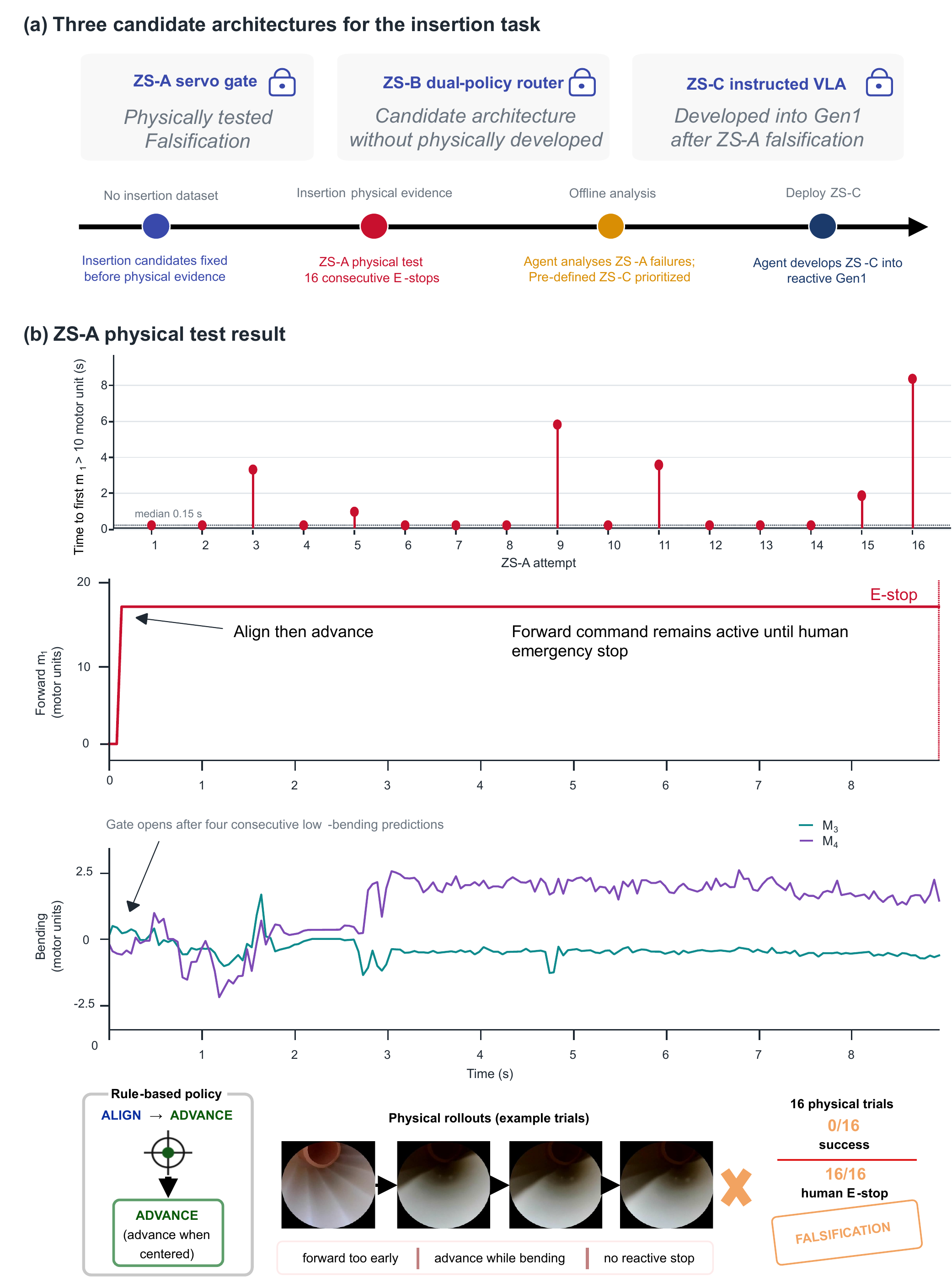}
\caption{\textbf{Physical experimentation falsifies an offline-supported insertion strategy.} \textbf{a}, Three pre-defined directions and their dispositions: ZS-A servo gate, dual-policy router (ZS-B) and instructed VLA policy (ZS-C). \textbf{b}, First-lunge times for all 16 ZS-A attempts, a representative forward-command and bending trace, and representative frames illustrating premature advancement without a reactive stop. All 16 attempts failed and required emergency stopping. The median first-lunge time was 0.15 s; the gate required four consecutive low-bending predictions.}
\label{fig:falsification}
\end{figure}

The agent operationalised the supported relations as zero-shot composition A (ZS-A): the tracking policy controlled bending, while four consecutive low-bending predictions opened a gate that issued a 16.7-unit forward command (Fig.~\ref{fig:falsification}). Offline replay appeared consistent with the intended logic, with the gate opening more often on low-error than high-error frames. ZS-A, a dual-policy router (zero-shot composition B, ZS-B) and an instructed VLA policy (zero-shot composition C, ZS-C) were defined before the agent observed any insertion trajectory or outcome.

Physical testing falsified ZS-A as a viable insertion policy. It completed 0/16 physical attempts (Wilson 95\% CI, 0--19.4\%), and all 16 attempts required emergency stopping. Every trial contained a single-tick forward increase of at least 10 motor units; the median maximum increase was 16.7 units and the median first-lunge time was 0.15~s. Testing ended after 16 consecutive emergency stops for safety. Source datasets had supported alignment, conditional advancement and the offline gate behaviour, yet they did not reveal that their discontinuous composition would lunge or that the composition lacked a contact-conditioned stop response.

Physical evidence therefore changed what the agent built. The source datasets did not contain the state transitions induced by deploying the discontinuous composition in closed loop. The dual-policy router was not deployed because it shared the staged composition and lacked an independently learned stop response. Investigators prioritised an end-to-end, stop-capable policy, and the agent developed the existing VLA candidate into Gen1. It paired features from the pretrained Qwen3-VL-4B vision--language model (VLM)~\citep{bai2025qwen3vl} with a flow-matching action head~\citep{lipman2023flow}, constructed alignment and advancement records from the source skills, added conservative stop proxies and implemented an independent positive forward-slew limit.

\subsection*{Physical evidence supports continued policy improvement}

\begin{figure}[!htbp]
\centering
\includegraphics[width=\linewidth,height=0.69\textheight,keepaspectratio,trim={527bp 0bp 0bp 0bp},clip]{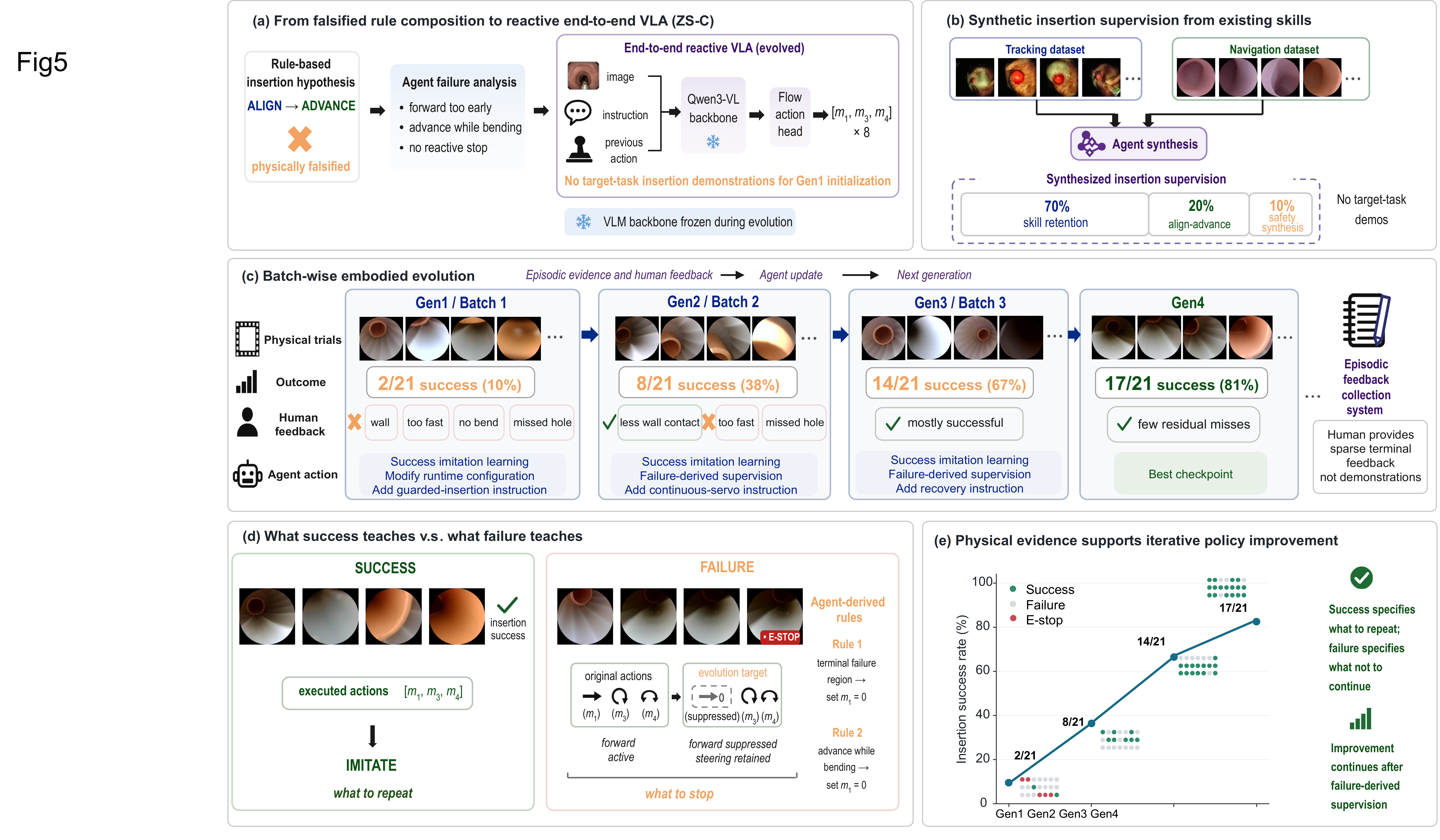}
\caption{\textbf{Physical evidence supports continued policy improvement.} \textbf{a}, Development of the instructed VLA direction into a reactive policy. \textbf{b}, Source-derived supervision used to initialise Gen1; the mixture contains 70\% skill retention, 20\% composed insertion and 10\% stop proxies. \textbf{c}, Physical outcomes and research revisions across Gen1--Gen4, including changes to instructions and execution settings. Gen1-to-Gen2 also included success-imitation action-head fine-tuning. \textbf{d}, Successful actions supply cloning targets; unsuccessful states supply no-forward targets with steering retained. \textbf{e}, Success rates across 21 attempts per generation; the line connects the observed proportions, and each dot denotes one trial outcome (S, F or E). Wilson 95\% CIs and exact outcomes are reported in Table~\ref{tab:phantom}; exact comparisons are in Supplementary Table~12. Generation comparisons include checkpoint, instruction and execution-profile changes and do not isolate their individual effects.}
\label{fig:evolution}
\end{figure}

Across four policy generations, physical insertion success increased progressively from 2/21 for Gen1 to 8/21 for Gen2, 14/21 for Gen3 and 17/21 for Gen4 (Fig.~\ref{fig:evolution}). The first update imitated the two successful Gen1 trajectories and used none of the unsuccessful trajectories. This success-only bootstrap was accompanied by a guarded-insertion instruction and a revised execution profile. Gen2--Gen4 shared the revised execution settings, while their task instructions progressed from guarded advancement to continuous visual servoing and recovery-oriented insertion (Methods and Supplementary Note~5).

Failed rollouts first contributed training targets in the Gen2-to-Gen3 update. The preceding batch contained eight successful and 13 unsuccessful physical outcomes. In eligible records, the agent converted states near unsuccessful terminations, and states combining forward motion with strong bending, into conservative avoidance targets: desired forward motion became zero while the recorded steering targets were retained. The revision therefore used failed behaviour to specify when advancement should cease, alongside successful actions that remained cloning targets. Gen3 subsequently completed 14/21 attempts, and the next combined update produced Gen4 at 17/21. By comparison, the Success-only control lineage, which repeatedly learned only from its own successful trials, plateaued at 3/21. Final Gen4 also exceeded the strongest autonomous reference, image-based visual servoing (IBVS), at 3/21; exact confidence intervals and two-sided Fisher comparisons are reported in Supplementary Tables~11 and 12.

This divergence is consistent with failed trajectories providing information about states in which forward motion should be suppressed. Because the two lineages differed in positive-evidence pools, checkpoints, instructions and execution settings, the comparison does not isolate a causal effect of failure supervision. It nevertheless shows that policy improvement continued after negative physical evidence entered the research loop, whereas repeated success-only updating did not produce the same progression. Generation-specific update recipes are reported in Supplementary Table~10.

\begin{table}[!htbp]
\centering
\caption{\textbf{Physical insertion outcomes in the benchtop development environment.} All arms contain 21 attempts except ZS-A, which stopped after 16 consecutive emergency stops. Success proportions use Wilson 95\% CIs; ordinary failures and emergency stops remain unsuccessful in every denominator. Exact two-sided Fisher comparisons are reported in Supplementary Table~12.}
\label{tab:phantom}
\small
\begin{tabularx}{\linewidth}{Yrrr}
\toprule
Policy arm & Success / $n$ & Success, \% (95\% CI) & Emergency stop \\
\midrule
ZS-A rule composition & 0/16 & 0.0 (0.0--19.4) & 16 \\
\midrule
CNN behaviour cloning & 0/21 & 0.0 (0.0--15.5) & 0 \\
Action-chunking transformer & 1/21 & 4.8 (0.8--22.7) & 1 \\
Diffusion Policy & 2/21 & 9.5 (2.7--28.9) & 0 \\
IBVS & 3/21 & 14.3 (5.0--34.6) & 0 \\
\midrule
Success-only, final iteration & 3/21 & 14.3 (5.0--34.6) & 0 \\
\midrule
Gen1 & 2/21 & 9.5 (2.7--28.9) & 5 \\
Gen2 & 8/21 & 38.1 (20.8--59.1) & 0 \\
Gen3 & 14/21 & 66.7 (45.4--82.8) & 0 \\
Gen4 & 17/21 & 81.0 (60.0--92.3) & 0 \\
\bottomrule
\end{tabularx}
\end{table}

\subsection*{Policy transfer across deployment shifts and tissue}

\begin{figure}[!htbp]
\centering
\includegraphics[width=\linewidth,height=0.86\textheight,keepaspectratio,trim={23.29bp 4.90bp 17.50bp 14.09bp},clip]{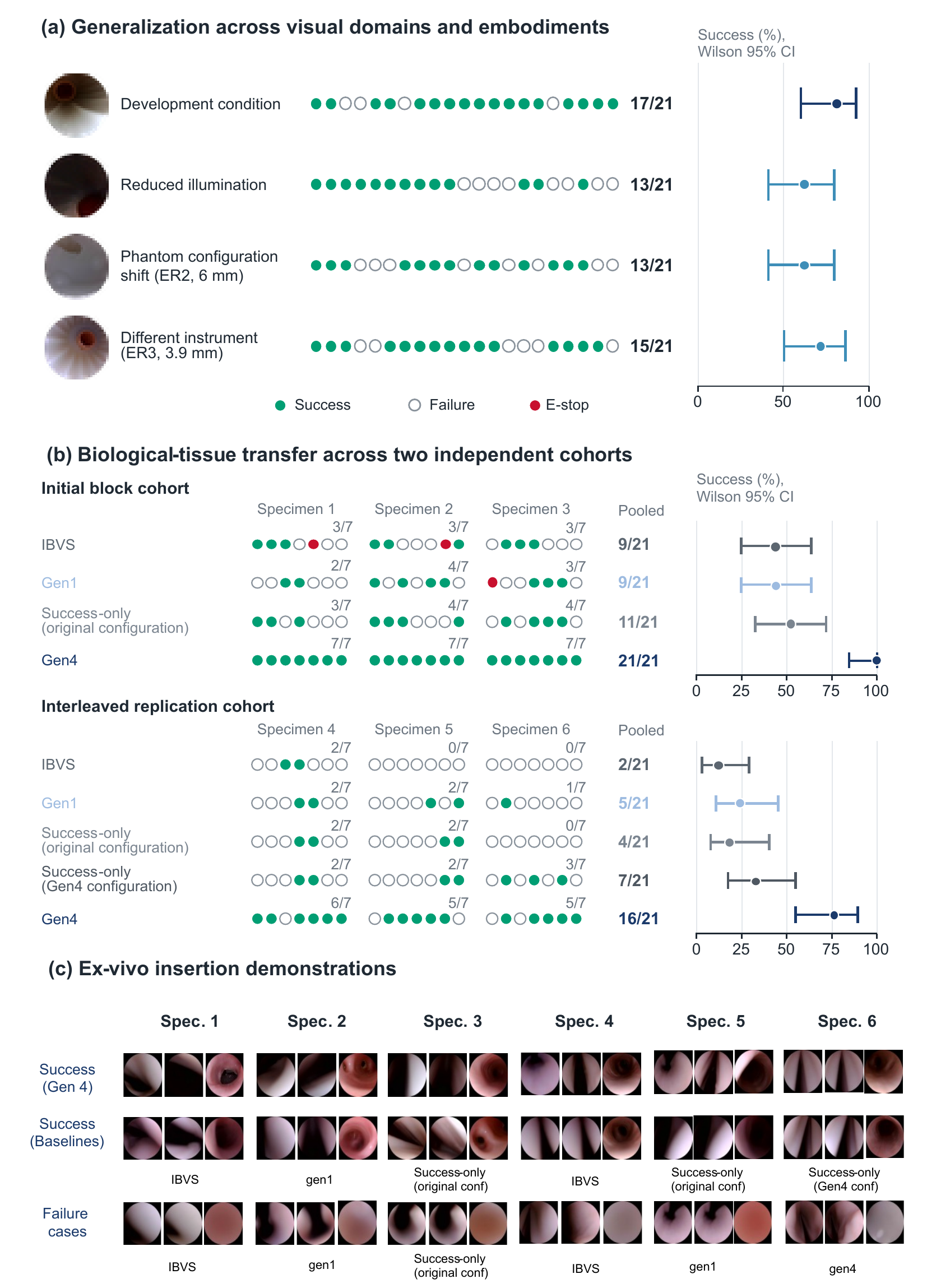}
\caption{\textbf{The evolved policy transfers across deployment shifts and biological tissue.} \textbf{a}, Gen4 outcomes across deployment conditions (21 attempts per condition). \textbf{b}, Autonomous outcomes in two ex-vivo cohorts (three specimens per cohort; seven attempts per condition and specimen). Dots show individual outcomes; intervals show Wilson 95\% confidence intervals. \textbf{c}, Representative starting, entrance-alignment and lumen or terminal views from successful Gen4, successful baseline and failed trials. Failure sequences do not establish lumen centring. These examples add no trials or biological replicates.}
\label{fig:transfer}
\end{figure}

After development, Gen4 was evaluated without adaptation under three deployment shifts (Fig.~\ref{fig:transfer}). It retained 13--15 successes in 21 attempts under reduced illumination, a phantom-configuration shift and a different motorised endoscope. Performance declined relative to the development condition but did not collapse, indicating that the learned capability was not confined to one visual appearance or instrument.

We next asked whether the policy evolved in the benchtop phantom retained competence in biological tissue. The Gen4 policy was evaluated in an initial block-design cohort of three independent ex-vivo porcine specimens together with the autonomous reference policies IBVS, Gen1 and Success-only. Gen4 was the highest-performing autonomous condition on each specimen, indicating that competence acquired during benchtop evolution was not restricted to the development phantom.

Generation-specific instruction profiles accompanied policy development, whereas the ex-vivo replication used a common task instruction across learned-policy conditions. We then evaluated this finding in a second independent cohort of three new specimens using an interleaved protocol and an expanded control set that included an execution-configuration control and human teleoperation references. Gen4 again remained the highest-performing autonomous policy on all three specimens and achieved 16/21 successful insertions, compared with 2--7/21 for the autonomous controls. Experienced and novice teleoperation references both achieved 21/21, showing that substantial headroom remained relative to human-controlled navigation. Applying the Gen4 execution configuration (runtime action scaling, gains and smoothing; Supplementary Note~8) to the Success-only checkpoint increased its success from 4/21 to 7/21 but did not close the gap to Gen4. Across the two cohorts, the six independent specimens were the biological units; repeated attempts were technical repetitions. Full specimen-level outcomes are reported in Supplementary Table~19. Figure~\ref{fig:transfer}b shows the autonomous conditions; the teleoperation references are included in the full source table. The two Success-only conditions used the same checkpoint and common task instruction, so their contrast tested the execution profile.

Representative successful and failed endoscopic sequences illustrate within-trial progression across the six specimens (Fig.~\ref{fig:transfer}c). Each triplet contains the recorded starting view, an entrance-alignment view and a later lumen or terminal view; the policy is labelled beneath each baseline or failure sequence. The selected alignment views can remain off-centre, and terminal failure views do not demonstrate successful lumen centring. These examples illustrate recorded trajectories without adding trials or biological replicates. Extended temporal sequences are shown in Supplementary Figs.~2 and 3.

\section*{Discussion}

EndoForge shows that physical robot experiments can function not only as performance evaluations, but as evidence that changes subsequent AI-driven policy development. Validation caused the agent to abandon an inferior action representation; physical deployment falsified a plausible source-skill composition; and failed rollouts altered the supervision used for later policies. The central result is therefore not the final success rate alone, but the demonstrated sequence in which physical evidence changed what the agent believed and built.

Traditional imitation learning, reinforcement learning and online adaptation optimise policies within representations, objectives and update rules specified beforehand~\citep{shafiullah2022bet,argall2009lfd,hussein2017imitation,oh2025rlalgorithms,schaul2016per,andrychowicz2017her}. EndoForge instead operates one level above that optimisation loop, using experimental evidence to decide whether representations, policy compositions and subsequent supervision should be revised. Generalist VLA policies provide a transferable controller substrate~\citep{bousmalis2024robocat,ghosh2024octo,bjorck2025groot,li2026robovlm}, while program-synthesis and multi-agent systems provide mechanisms for grounded planning, software generation and coordination~\citep{singh2023progprompt,huang2022zeroshot,ji2026genswarm,monwilliams2025,mower2026rosllm,carvalho2026iterative,xiao2026enpire}. AI-scientist systems and self-driving laboratories connect hypotheses to experiments in digital workflows or structured apparatus~\citep{lu2026aiscientist,macleod2022selfdriving,burger2020robotchemist,king2004robotscientist}. Physical robots add a distinct challenge because a policy changes the states it subsequently observes, and a seemingly supported hypothesis can fail through contact, delay or closed-loop state transitions that are absent from source data.

These observations suggest that physical research agents should be evaluated by more than their final task success. Relevant questions include whether a candidate was fixed before an experiment, whether the physical result could reject it, whether that evidence changed the next implementation, whether failed trials contributed information unavailable from successes and whether the resulting policy transferred beyond its development conditions. Together, these criteria distinguish an evidence-responsive research loop from ordinary repeated model fitting.

The ZS-A result provides the clearest physical falsification. Human investigators supplied an align-then-advance task decomposition, and the agent found quantitative support for its components. Offline replay further confirmed that the implemented gate behaved as designed. Although the composition was supported by source-task statistics and offline replay, physical testing falsified it by exposing a hazardous closed-loop behaviour absent from those data. This distinction---between implementing a reasonable hypothesis and allowing embodiment to reject it---is central to the research capability tested here.

The generational sequence provides a second form of evidence. Successful trajectories were sufficient to bootstrap Gen2, but continued gains appeared after hazardous terminal states were converted into avoidance targets. The plateau of the Success-only control is compatible with failed trajectories expanding the information available for revision. However, instruction changes and the Gen1-to-Gen2 execution-profile change accompanied development, so the generation series cannot separate their effects from those of action-head updates or failure supervision.

The final policy also retained competence across visual and hardware shifts and transferred without adaptation to two ex-vivo cohorts. Gen4 was the highest-performing autonomous condition on every discarded porcine specimen. Importantly, its relative advantage was reproduced in three additional specimens under an interleaved protocol with expanded autonomous and teleoperation controls. Nevertheless, the 16/21 replication result and the perfect teleoperation references leave a clear gap between current autonomy and human-controlled navigation. The tissue lumen narrowed progressively towards the opening, unlike the planar benchtop aperture, which may have permitted contact-aided centring after small residual misalignment; this interpretation remains a geometry-specific hypothesis rather than a demonstrated general mechanism.

The execution-configuration control further showed that the final gap was not merely a calibration artefact: Gen4's runtime settings recovered part, but not all, of the difference when applied to the Success-only checkpoint.

EndoForge is not a fully autonomous scientist. Human investigators defined the research goal, supplied the initial align-then-advance hypothesis, controlled deployment and safety, and provided terminal outcomes. The framework automates selected stages of robot-policy research: analysing evidence, implementing candidates, interpreting failures and constructing subsequent updates. Its present evidence is further limited to one endoscopic insertion task, six biological specimens and single-operator teleoperation references; it does not establish population-level operator effects or a causal estimate of failure supervision.

Future studies should test the framework across prospectively randomised policy evaluations, multiple operators, robot systems and tasks. More broadly, the study suggests a design pattern for safety-bounded physical robot-policy research: separate research-time agency from control-time authority, expose policy hypotheses to physical experiments and require experimental evidence to change subsequent development.

\FloatBarrier

\section*{Methods}

\subsection*{EndoForge research loop and coding-agent implementation}

Within EndoRSI, EndoForge organised policy development as a sequence of evidence analysis, candidate implementation, authorised physical evaluation and revision. The research-time coding agent used Qwen3.7-Max~\citep{qwen2026qwen37}. Its workspace contained source image--action datasets, policy and analysis code, validation outputs and records of completed physical trials. These records linked recorded observations and executed actions to terminal success, failure or emergency-stop outcomes. Investigators defined the research objective and experimental constraints, proposed the initial align-then-advance decomposition, and determined when a candidate could be deployed.

Before physical testing, the agent audited the source datasets, implemented reference and candidate policies, and used validation results to retain or reject candidate designs. It then analysed relationships between source-task alignment, bending and advancement to construct insertion candidates. Subsequent physical evidence informed changes to policy implementation, training-target construction, checkpoint weights and deployment instructions. Outputs included candidate code, analysis results, checkpoints and manifests identifying the policy and execution configuration to be evaluated.

Each physical evaluation used one fixed candidate. Human authorisation separated candidate preparation from execution: the coding agent could revise future candidates between experiments but had no live camera, serial or motor interface while the robot was armed. The operator restored the initial configuration, authorised motion, assigned the terminal outcome and retained emergency-stop control. Completed-trial evidence then returned to the coding workspace for analysis before the next candidate was prepared. The recorded lifecycle and decision chronology are documented in Supplementary Notes~1 and 7.

The study comprised 331 benchtop attempts and 231 ex-vivo attempts. Policy generations were successive development batches, not randomised parallel treatment arms. The biological-tissue evaluation used six independent tracheal specimens that were not part of policy development. These design features determine the units of analysis and the scope of the comparisons described below.

\subsection*{Source datasets and offline policy development}

Source tracking data were collected with a medical endoscope in discarded ex-vivo porcine stomach tissue, and source navigation data with an industrial endoscope in a colon phantom. Each task comprised 100 episodes. A fixed split assigned 80 episodes to training, 10 to validation and 10 to held-out testing, with all frames from an episode confined to one partition. Actions were expressed in semantic insertion and bending coordinates, $[m_1,m_3,m_4]$. Because the source platforms operated at different motor scales, each task retained its own action normalisation with the checkpoint.

Image-conditioned references included CNN behaviour cloning, Action Chunking with Transformers, Diffusion Policy and a vision--language--action (VLA) policy. Evaluation used motor-unit mean absolute error (MAE), root-mean-square error and sign accuracy. A separate copy-previous diagnostic received the preceding ground-truth action to assess temporal autocorrelation; it was not an autonomous visual controller. The agent's first candidate represented future actions as discrete bins plus continuous residuals. After validation error increased, the agent replaced this representation with direct continuous prediction from the current image and temporal differences. The replacement predicted eight-step action chunks, and causal temporal ensembling combined overlapping predictions. Held-out comparisons and three independent training runs followed design selection (Supplementary Tables~5, 6, 7 and 8).

\subsection*{Reactive policy synthesis and failure-informed updates}

The initial insertion candidates combined the two source skills without target-task insertion demonstrations. ZS-A used tracking-based steering and a gate that initiated advancement after sustained low bending. A dual-policy router and an instructed VLA candidate were fixed alongside ZS-A before insertion testing. Repeated unsafe advancement during ZS-A testing prompted investigators to prioritise a stop-capable end-to-end policy; the agent developed the existing VLA direction, and the router was not physically deployed.

The reactive policy combined pretrained Qwen3-VL-4B features~\citep{bai2025qwen3vl} with a flow-matching head~\citep{lipman2023flow} predicting eight future semantic actions. Source tracking supplied alignment records, low-bending navigation supplied advancement records, and close-target tracking, navigation stop onset and stopped hard-correction states supplied conservative stop proxies. The training mixture retained source-skill records alongside composed insertion and stop records. Thus, the initial insertion policy was trained from existing skills and derived supervision before receiving physical insertion experience.

Following physical evaluation, successful trajectories supplied action-cloning targets. The update producing Gen2 used only successful trajectories. Updates producing Gen3 and Gen4 additionally constructed avoidance targets from unsuccessful trajectories: forward targets were set to zero in the terminal portion of a failed trajectory and in states combining advancement with large bending demand, while the recorded bending targets were retained. This supplied examples of when advancement should cease without prescribing a new steering direction. Only records passing frame/action and action-availability checks contributed optimisation samples; unsuccessful attempts remained in the outcome denominator regardless of training eligibility. These embodied updates trained the action head while keeping the vision--language backbone and action normalisation fixed. Optional human free text changed no training sample.

Development also involved revised deployment instructions and execution profiles. Gen1 used the original profile, whereas Gen2--Gen4 shared a revised profile with different action scaling and smoothing. The separate Success-only lineage learned from successes in its own preceding batch and discarded unsuccessful trajectories. It therefore did not share a matched trajectory pool or successor checkpoint with the generation lineage. In ex-vivo replication, all learned policies received a common instruction, and a configuration control applied the Gen4 execution profile to the final Success-only checkpoint. Exact synthesis rules, target equations, prompts and update recipes are specified together in Supplementary Note~5.

\subsection*{Robotic platforms and execution}

Endoscopic robot~1 (ER1) used a 9.9-mm Olympus GIF-Q260J medical gastroscope for source tracking. ER2 used a 6.0-mm Ailishi S50 industrial endoscope for source navigation, benchtop policy development and ex-vivo evaluation. ER3 used a 3.9-mm Ailishi S50 endoscope for the different-instrument transfer arm. Platform adapters translated the shared semantic action into instrument-specific motor commands.

Physical insertion used a 400$\times$400 RGB image stream and a separate 20-Hz execution process. Policy inference ran asynchronously; the runtime consumed the newest valid action chunk and issued zero action when no valid plan was available or the watchdog expired. Fixed magnitude limits and a positive forward-slew limit bounded executed commands. Human emergency-stop control overrode motor output.

\subsection*{Benchtop and deployment-shift evaluation}

ER2 performed autonomous insertion through a circular 10-mm aperture in a white silicone phantom. Each attempt began with the distal tip 5~cm from the aperture and a random lateral offset that kept the opening visible but away from the image centre. The apparatus was reset between attempts. During autonomous execution, the operator supplied no steering command and intervened only to terminate an attempt or activate the emergency stop.

Success required passage of the distal tip through the aperture plane before termination and without an emergency stop. Ordinary failure denoted an unsuccessful attempt terminated for manual takeover; an emergency stop denoted immediate termination of unsafe behaviour. Both were unsuccessful outcomes. Terminal outcomes were assigned by the operator. The archived review runtime includes a configurable motor-zeroing timeout, set to 180~s by default; timeout expiry stops motion but does not itself close the trial or assign an outcome (Supplementary Note~6). Each benchtop arm comprised 21 attempts, except ZS-A, which was halted after 16 consecutive emergency stops for safety; all 16 attempts remained in the analysis.

The final Gen4 checkpoint was also evaluated without adaptation in three 21-attempt deployment-shift arms. Reduced illumination switched off the endoscope's distal light while leaving ambient lighting unchanged. The phantom-configuration shift used a smaller 7-mm aperture beside a protruding distractor, with the same silicone material and background. Different-instrument transfer replaced ER2 with ER3 while retaining the task and checkpoint. The original Gen4 generation experiment supplied the development-condition reference.

\subsection*{Ex-vivo evaluation}

Fixed policies were evaluated on six discarded porcine tracheae in two independent cohorts of three specimens. The initial cohort compared Gen4, Gen1, IBVS and final Success-only in condition blocks. The replication cohort interleaved these autonomous conditions with the Success-only execution-configuration control and experienced and novice teleoperation references. Each condition comprised seven attempts per specimen, yielding 84 initial-cohort and 147 replication attempts. There was one teleoperator in each experience category.

Success required traversal of the distal tip from the specimen-specific starting end to the predefined target lumen at the opposite end. Policy state was reset between attempts, and the same success, ordinary-failure and emergency-stop categories were used as in benchtop testing. Specimens were independent biological units; repeated attempts within a specimen were technical repetitions. The cohorts retained separate analyses because their condition sets and ordering protocols differed. The recorded schedules, session timing and deviations from the intended interleaving are documented in Supplementary Note~8.

\subsection*{Outcome assessment and statistical analysis}

Live terminal outcomes supplied physical-outcome supervision. Independent video review assessed binary outcomes and successful ex-vivo navigation times. Binary agreement across three sources was 100\% for all 546 scored trials, including all 84 initial-cohort ex-vivo trials (Supplementary Table~17). Emergency-stop counts used live records because video alone could not establish activation. Successful-navigation times measured traversal between tissue ends and used the median of available exact-path reviewer annotations. They were conditional on success and were not raw video durations.

Binary proportions used two-sided Wilson 95\% confidence intervals~\citep{wilson1927probable}; specified two-arm comparisons used two-sided Fisher exact tests~\citep{fisher1922interpretation}. Offline uncertainty used paired bootstrap resampling of held-out episodes~\citep{efron1979bootstrap}, with episodes weighted equally. Frames were not independent replicates. Ex-vivo pooled attempt-level comparisons were exploratory and did not provide specimen-level inference. Generation comparisons describe observed outcomes across changing checkpoints, instructions and execution profiles; they do not isolate the causal effect of failure supervision. Single-operator teleoperation references do not establish population differences between experience groups. No a priori power calculation was performed. Resampling settings, rating-reliability estimators and record-matching checks are specified in Supplementary Note~9.

\subsection*{Ethics and tissue sourcing}

No live animals were used or killed for this study. The ex-vivo porcine stomach used for source tracking and the porcine tracheae used for transfer were discarded post-slaughter tissues obtained from the same vendor at stall M36, Tai Po Hui Market, Tai Po, Hong Kong. The study involved no human tissue, patients or clinical decisions. The two teleoperation conditions were descriptive system-reference arms, with one operator in each experience category.

\subsection*{AI-assisted research and writing disclosure}

The Qwen3.7-Max coding-agent component is part of the scientific method. Generative-AI assistance was also used for language editing and structural suggestions. The authors checked trial records, statistics, references and scientific claims and retain responsibility for the manuscript.

\begingroup\small
\setlength{\bibsep}{1pt}

\endgroup

\begingroup
\singlespacing
\section*{Acknowledgements}

We thank Yupeng Wang and Zhiqing Tang for the original hardware prototypes. We also thank Xuran Chen for helping C.K.N. collect the ex-vivo tracking dataset.

This work was supported by the Ministry of Science and Technology (MOST) of China Key Projects (2025YFE0122500 and 2024YFE0216200); the Innovation and Technology Fund (ITF) of Hong Kong SAR (MHP/185/24); and the Hong Kong Research Grants Council (RGC) through the General Research Fund (GRF 14216022, 14204524, 14203323 and 14206125), the Research Impact Fund (RIF R4020-22) and the Strategic Topics Grant (STG 1/M-405/25-N).

\section*{Author Contributions}

C.K.N. and Y.Z. contributed equally to this work. C.K.N. conceived the study, developed the methodology and software, coordinated the project, performed experiments and investigations, conducted formal analyses, prepared the visualizations, and wrote the original manuscript. C.K.N. also led the project and contributed to manuscript review and editing. Y.Z. contributed to software development, experimental investigation, data collection, and project coordination. T.W. contributed to software development and experimental investigation. H.Z. contributed to software development and experimental investigation. S.H.L. designed and fabricated the ER2 and ER3 robotic hardware platforms. H.G. designed and fabricated the ER1 robotic hardware platform. H.R. provided resources and supervision, acquired funding, and contributed to manuscript review and editing. All authors reviewed and approved the final manuscript.

\section*{Competing Interests}

The authors declare no competing interests.
\endgroup

\clearpage
\newgeometry{margin=0.82in}
\singlespacing
\setcounter{figure}{0}
\setcounter{table}{0}
\setcounter{equation}{0}
\renewcommand{\figurename}{Supplementary Figure}
\renewcommand{\tablename}{Supplementary Table}
\renewcommand{\theHfigure}{supp.\arabic{figure}}
\renewcommand{\theHtable}{supp.\arabic{table}}
\renewcommand{\theHequation}{supp.\arabic{equation}}
\captionsetup{font=small,labelfont=bf,labelsep=period}
\pdfbookmark[0]{Supplementary Information}{supplementary-information}
\begin{center}
{\Large\bfseries Supplementary Information\par}
\vspace{0.6em}
{\large\bfseries EndoRSI: Physical-Evidence-Driven Recursive Improvement of Endoscopic Robot Policies
\par}
\end{center}
\section*{Contents}
\begingroup
\singlespacing
\small
\setlength{\parindent}{0pt}
\newcommand{\siindex}[2]{\noindent\hyperref[#2]{#1}\nobreak\dotfill\nobreak\pageref{#2}\par\vspace{0.45em}}
\subsection*{Supplementary Notes}
\siindex{Note 1. Evidence-to-policy lifecycle and authority boundary}{si-note:1}
\siindex{Note 2. Robot platforms and software architecture}{si-note:2}
\siindex{Note 3. Offline model selection}{si-note:3}
\siindex{Note 4. Cross-task analysis and ZS-A implementation}{si-note:4}
\siindex{Note 5. Reactive VLA synthesis and embodied updates}{si-note:5}
\siindex{Note 6. Benchtop experimental protocol}{si-note:6}
\siindex{Note 7. Human--agent decision chronology}{si-note:7}
\siindex{Note 8. Ex-vivo protocol and outcome assessment}{si-note:8}
\siindex{Note 9. Statistical analysis and rating reliability}{si-note:9}
\subsection*{Supplementary Figures}
\siindex{Figure 1. Endoscopes and robot configurations}{si-fig:robots}
\siindex{Figure 2. Ex-vivo evaluation and a successful Gen4 trial}{si-fig:exvivoqual}
\siindex{Figure 3. Representative unsuccessful ex-vivo trials}{si-fig:exvivofail}
\subsection*{Supplementary References}
\siindex{References}{si-si:references}
\clearpage\section*{Contents continued}
\subsection*{Supplementary Tables}
\siindex{Table 1. Implemented evidence-to-policy lifecycle}{si-tab:lifecycle}
\siindex{Table 2. Control flow and safety authority}{si-tab:controlpath}
\siindex{Table 3. Instruments, experiments and controllers}{si-tab:platforms}
\siindex{Table 4. Collection, policy and hardware code}{si-tab:codepath}
\siindex{Table 5. Source datasets and experimental units}{si-tab:datasets}
\siindex{Table 6. Model selection and held-out performance}{si-tab:offlineselection}
\siindex{Table 7. Complete held-out offline results}{si-tab:offline}
\siindex{Table 8. Stability across three training runs}{si-tab:seeds}
\siindex{Table 9. Source-task evidence and disposition}{si-tab:cross}
\siindex{Table 10. Generation-specific update recipes}{si-tab:trainingrules}
\siindex{Table 11. Complete benchtop outcomes}{si-tab:phantomfull}
\siindex{Table 12. Fisher tests for phantom contrasts}{si-tab:phantomstats}
\siindex{Table 13. Human and agent scientific decisions}{si-tab:provenance}
\siindex{Table 14. Initial ex-vivo outcomes by specimen}{si-tab:exvivofull}
\siindex{Table 15. Cohort-specific descriptive comparisons}{si-tab:exvivosensitivity}
\siindex{Table 16. Trial-video duration phenotypes}{si-tab:durationphenotypes}
\siindex{Table 17. Binary outcome agreement}{si-tab:rateragreement}
\siindex{Table 18. Navigation-time annotation reliability}{si-tab:timingreliability}
\siindex{Table 19. Transfer outcomes across six specimens}{si-tab:exvivocombined}
\endgroup\clearpage

\section*{Supplementary Note 1: Evidence-to-policy lifecycle and authority boundary}\label{si-note:1}

Within EndoRSI, the EndoForge research-time coding agent used Qwen3.7-Max; the reactive robot policy separately used Qwen3-VL-4B. The coding workspace contained source image--action records, policy and analysis code, validation outputs and completed-trial evidence. These inputs supported dataset audit, candidate implementation, offline comparison and subsequent revisions. Supplementary Table~\ref{si-tab:lifecycle} identifies the evidence and retained output at each research stage.

A candidate consisted of the policy implementation and checkpoint together with its manifest, task instruction and execution configuration. Investigators defined the task and experimental constraints and authorised physical deployment. While a trial was armed, the candidate remained fixed and the coding agent had no live camera, serial, motor or emergency-stop interface. The runtime translated policy predictions into bounded motor commands, and the operator retained reset, termination and emergency-stop authority. Supplementary Table~\ref{si-tab:controlpath} specifies these interfaces.

After a trial, recorded observations, actions and the human-assigned success (S), ordinary-failure (F) or emergency-stop (E) outcome became available for subsequent analysis. Research outputs included candidate code, analysis results, selected checkpoints and generation-specific recipes. Human investigators supplied the initial align-then-advance decomposition but no insertion motor targets or physical insertion demonstrations. The ordered decisions actually recorded during development are presented separately in Supplementary Note~7.

\section*{Supplementary Note 2: Robot platforms and software architecture}\label{si-note:2}

\begin{figure}[!htbp]
\centering
\includegraphics[width=0.98\linewidth]{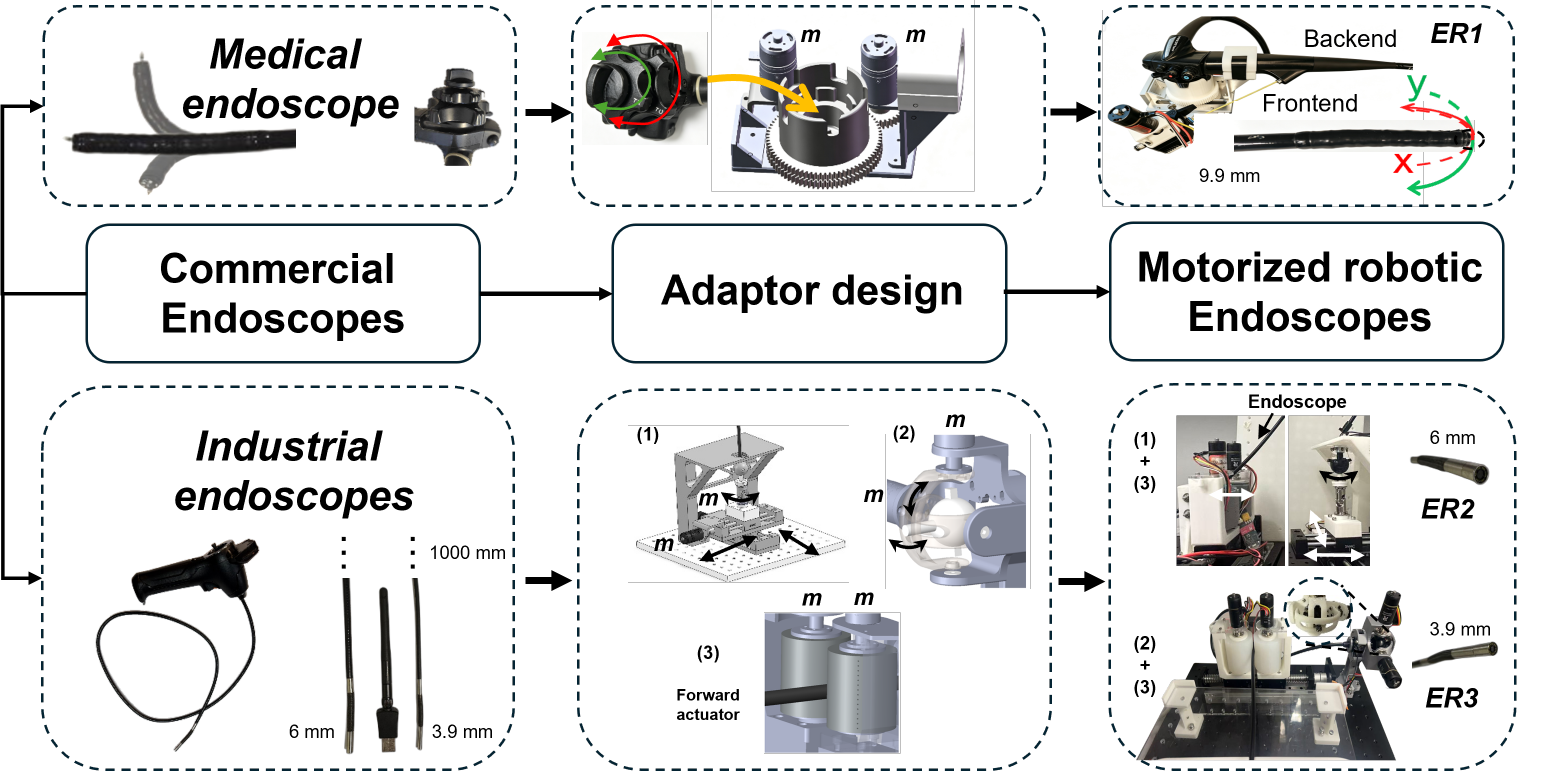}
\caption{\textbf{Commercial endoscopes and motorised robot configurations.} The upper row shows the medical gastroscope, its motorised knob adapter and the assembled ER1 platform (Olympus GIF-Q260J; 9.9-mm insertion diameter). The lower row shows industrial endoscopes and adapter designs used for ER2 (Ailishi S50; 6.0 mm) and ER3 (Ailishi S50; 3.9 mm). Labels (1) and (2) identify the illustrated industrial steering adapters; (3) identifies the shared forward actuator. ER1 supplied tracking data, ER2 supplied navigation data and supported policy development and ex-vivo experiments, and ER3 was used for different-instrument transfer. The original hardware illustration is reproduced without modification. Dimensions are those labelled in the supplied illustration; the arrangement is not a calibrated scale comparison.}
\label{si-fig:robots}
\end{figure}

The hardware illustration shows how commercial instruments were motorised using platform-specific adapters. The three motorised endoscopes share one semantic policy interface but not one mechanical or firmware implementation (Supplementary Fig.~\ref{si-fig:robots} and Supplementary Table~\ref{si-tab:platforms}). ER1 is the medical-endoscope platform used to collect tracking data. ER2 is the 6.0-mm industrial platform used to collect navigation data and to conduct benchtop development, evolution, the low-illumination and phantom-configuration shifts, and ex-vivo experiments. ER3 is the 3.9-mm industrial platform used for the different-instrument transfer arm. The 6.0-mm and 3.9-mm systems share the same six-channel lower-controller protocol, whereas ER1 uses a separate four-motor firmware. The physical insertion controller received 400$\times$400 red--green--blue (RGB) frames at 20~Hz; policy-training inputs were resized to 224$\times$224.

Data collection, training, inference and actuation were kept as explicit software layers. Each platform-specific collection program stored the camera stream, timestamps and executed motor state. Dataset conversion aligned images with actions and expressed every valid training record in the canonical semantic coordinates $[m_1,m_3,m_4]$ before episode splitting and task-specific normalisation. Training therefore consumed device-independent records rather than raw serial packets. At inference, the image-conditioned controller produced semantic action chunks at the host; a 20-Hz process selected the newest valid action, enforced the watchdog and limits, and passed the command to a platform adapter. For ER2 and ER3, the adapter expanded $m_1$ to paired insertion motors M1/M2, applied the configured signs, gains and biases, forwarded $m_3,m_4$ to the two bending axes and set M5/M6 to zero. ER1 used its own four-motor packet mapping. The lower firmware then converted the packet to physical motor actuation. This separation is what allows training data from heterogeneous robots to share a policy representation while preserving device-specific mechanics at deployment.

Multimodal inference ran asynchronously. The semantic action $[m_1,m_3,m_4]$ represented forward/retract motion and two orthogonal bending commands. The runtime emitted zero action before a valid plan, after planner underrun or after a 150-ms watchdog expiry.

Absolute limits were $|m_1|\leq42$ and $|m_3|,|m_4|\leq120$ motor units. Reactive insertion releases limited positive changes in forward magnitude to 2.0 units per 50-ms tick.

\section*{Supplementary Note 3: Offline model selection}\label{si-note:3}

\subsection*{Source datasets and baseline evaluation}

Tracking used the Olympus GIF-Q260J medical endoscope robot (ER1) in an ex-vivo pig stomach; navigation used the Ailishi S50 6.0-mm industrial motorised endoscope (ER2) in a colon phantom. Each task contained 100 episodes. A fixed seed assigned 80 episodes to training, 10 to validation and 10 to held-out testing; no frame crossed episode partitions. Navigation contained 16,534 action rows and tracking contained 12,710 aligned image--action rows. Task-specific normalisation was used because tracking bending was approximately $[-9,10]$, whereas navigation spanned approximately $[-119,120]$ motor units. The canonical training record linked each image and timestamp to semantic $[m_1,m_3,m_4]$. Full source-dataset accounting is provided in Supplementary Table~5. The heterogeneous source systems reflect the range of robotic endoscopy platforms and learned endoscopic representations considered in recent work~\citep{si-kume2026acrs,si-slawinski2018magnetic,si-winters2022robotic,si-ng2025endovla,si-lin2026bilivla,si-lin2026endowam}.

Fixed image-conditioned references were CNN behaviour cloning with a ResNet-18 encoder~\citep{si-he2016resnet}, Action Chunking with Transformers~\citep{si-zhao2023act}, Diffusion Policy~\citep{si-chi2023diffusion}, classical IBVS~\citep{si-chaumette2006visual1,si-hutchinson2007visual2} and an image-conditioned VLA policy~\citep{si-kim2025openvla}. Comparisons used raw motor-unit MAE, root-mean-square error and sign accuracy. A copy-previous diagnostic supplied the preceding ground-truth action and was used only to expose temporal autocorrelation. Complete held-out comparisons are reported in Supplementary Table~7 and Supplementary Table~6.

\subsection*{Validation-guided candidate revision}

The version 1 (v1) candidate concatenated three RGB frames at the encoder input and predicted eight future $[m_1,m_3,m_4]$ actions. Each action dimension was represented by one of 11 bins plus a continuous residual. Its validation MAE was 2.654 for tracking and 19.811 for navigation, versus 2.619 and 19.463 for CNN behaviour cloning. The representation was rejected before the held-out comparison. The v1 held-out values in Supplementary Table~\ref{si-tab:offline} were calculated post hoc and were not used for model selection. Although its post hoc tracking MAE was low, its tracking RMSE and sign accuracy were worse than the CNN reference, and it showed no advantage on the navigation metrics.

The version 2 (v2) candidate used the current image and a temporal-difference image, concatenated as six channels. A ResNet-18 encoder fed direct $\ell_1$ regression of an $8\times3$ chunk. Exponential moving-average weights used decay 0.999. The v2+TE variant causally ensembled overlapping chunks:
\begin{equation}
\hat{a}_t=\frac{\sum_{j=0}^{J_t-1}w_j a^{(t-j)}_j}{\sum_{j=0}^{J_t-1}w_j},
\end{equation}
where $a^{(t-j)}_j$ is the action for time $t$ predicted $j$ ticks earlier and $w_j$ decreases with age. The selected design reduced validation MAE to 2.426 for tracking and 18.791 for navigation.

Three independent training runs were evaluated after design selection (Supplementary Table~8). Validation records were organised by episode; frames were never treated as independent replicates.

\section*{Supplementary Note 4: Cross-task analysis and ZS-A implementation}\label{si-note:4}

Human investigators described insertion as combining tracking-based alignment with navigation-based advancement. The agent converted this task decomposition into quantitative, falsifiable candidate designs. Tracking target-centre error was defined by $e_x=0.5-c_x$ and $e_y=0.5-c_y$, where $(c_x,c_y)$ is the normalised target-box centre. Least-squares fits over 10,284 training frames gave $m_3\simeq23.803e_y$ and $m_4\simeq32.325e_x$, with coefficients of determination $R^2=0.340$ and 0.370. An additional agent-tested navigation hypothesis, bending toward the centroid of a thresholded dark lumen, was unsupported ($R^2=-0.034$ and 0.004) and rejected.

Across 13,810 navigation training frames, advancement was grouped by $b=\max(|m_3|,|m_4|)$. Mean forward command was 16.746 motor units in the lowest-bending quartile and 0.560, 0.323 and 0.162 in the remaining quartiles. ZS-A waited for four consecutive normalised bending magnitudes below 0.35 and then issued $m_1=16.7$. A dual-policy router and an instructed VLA policy were fixed alongside ZS-A before any insertion evidence. Physical ZS-A testing produced 0/16 successes and 16 emergency stops. Because the dual-policy router inherited staged skill composition without an independently learned stop response, it was not physically deployed after the ZS-A result redirected development towards the end-to-end VLA policy. Every ZS-A trial contained a single-tick forward increase of at least 10 motor units; the median maximum increase was 16.7 and the median first-lunge time was 0.15~s.

The pre-frozen instructed VLA candidate used the unseen instruction ``insert into the hole''.

\section*{Supplementary Note 5: Reactive VLA synthesis and embodied updates}\label{si-note:5}

After ZS-A physical falsification and a human requirement for end-to-end stop-capable behaviour, the agent developed its pre-frozen VLA direction. Insertion synthesis used source tracking and navigation only. Tracking supplied alignment examples; low-bending navigation windows supplied advancement; close-target tracking, navigation stop onset and stopped hard-correction states supplied conservative stop targets. The effective mixture was 70\% retained source skills, 20\% composed insertion and 10\% stop proxies. Gen1 used features from the pretrained Qwen3-VL-4B vision--language model (VLM) and a flow-matching head~\citep{si-lipman2023flow} for eight future actions.

The converter expanded insertion instructions with five English and five Chinese phrasings. None of these records was a target-task insertion demonstration. Training used 224$\times$224 images, batch size 4, gradient accumulation 4, 12,000 steps, action-head learning rate $10^{-4}$ and learning rate $2\times10^{-6}$ for the final two VLM layers. An internal ablation with a higher stop-proxy sampling ratio did not improve stop recall and increased predicted lunge, so the agent retained the earlier release and left further correction to physical evidence.

The Gen1-to-Gen2 update was success only: two successful trials produced 175 imitation samples, while none of the 19 unsuccessful physical trials was used for training. Failure supervision first entered the Gen2-to-Gen3 update. Its source batch contained eight successful and 13 unsuccessful physical outcomes. After record-level frame/action and action-availability checks, 12 trial records contributed samples: two success-labelled records supplied cloning targets and ten failure- or emergency-stop-labelled records supplied avoidance targets. Nine records supplied no parameter-update sample. A pre-mirroring cap retained 72 avoidance examples; left--right mirroring produced 144 avoidance samples, which were combined with 68 cloning samples for a 212-sample, 300-step update. These training-eligibility fields do not redefine the 21 physical-outcome trials. VLM weights and normalisation remained fixed, and action loss decreased from 0.2243 to 0.0918.

The improvement from Gen1 to Gen2 and the smaller change in the Success-only control are not duplicate executions of one update. The generation update learned from the two successful trajectories in the Gen1 batch. The separate control learned only from successes in its own preceding physical batch, which began with one success. Thus, ``success only'' defines the evidence-selection rule, whereas the trajectory set and successor checkpoint are lineage specific. The control sequence of 1/21, 3/21, 3/21 and 3/21 tests whether repeated success-only updating reproduced the failure-informed generation trajectory.

For a failed trajectory with $T$ valid steps and normalised bending magnitude $b_t$, the evolution target was
\begin{equation}
\tilde{a}_t=
\begin{cases}
(0,m_{3,t},m_{4,t}), & t\geq0.65T,\\
(0,m_{3,t},m_{4,t}), & m_{1,t}>0\ \land\ b_t>0.35,\\
a_t, & \text{otherwise}.
\end{cases}
\end{equation}
The same rule also triggered when absolute bending exceeded 30 motor counts. Optional text could in principle enlarge a terminal stop region, but the two text-tagged records contributed no training sample and therefore changed no target. Text never generated a bending target. The update used 300 optimisation steps, batch size 4 and action-head learning rate $10^{-4}$, while freezing the VLM and retaining normalisation.

The Gen3-to-Gen4 source batch comprised 14 successful and seven unsuccessful trajectories; all 21 trajectories contributed to the update. It used 240 cloning and 160 avoidance samples, 300 steps, batch size 4 and learning rate $10^{-4}$; action loss decreased from 0.0914 to 0.0643. Each policy generation was evaluated using the success/failure/emergency-stop (S/F/E) criterion defined in Supplementary Note~6.

\subsection*{Task instructions and execution profiles during development}
Generation-specific instruction profiles accompanied policy development, whereas the ex-vivo replication used a common task instruction across learned-policy conditions. The exact deployment instructions were:
\begin{description}
\item[Gen1 and Success-only:] ``align to the target, then advance when centered''
\item[Gen2:] ``Perform guarded lumen insertion: use small steering corrections to reduce lateral and vertical error, never push while visibly off-center, advance in short controlled increments after stable alignment, and pause to re-center whenever drift appears.''
\item[Gen3:] ``Insert through the lumen with continuous visual servoing: keep the opening near image center, slow down as the rim approaches, re-align instead of forcing contact, and stop forward motion whenever bending demand is large or the target disappears.''
\item[Gen4:] ``Complete a safe, smooth lumen insertion under appearance and pose variation: acquire and center the opening, confirm stable alignment, advance progressively while preserving clearance, recover from drift with gentle corrections, and stop rather than push on uncertain alignment or resistance.''
\end{description}
Gen1 and the original-configuration Success-only control used tracking-scale bending, forward gain 1, bending gains 1 and 1, zero bending biases and no exponential smoothing. Gen2--Gen4 used navigation-scale bending, forward gain 1.3, bending gains 1.5 and 2.5, bending biases $-0.048$ and $0.111$, and a 10-s exponential-smoothing time constant. This runtime scaling is distinct from the fixed training normalisation. The Gen1-to-Gen2 transition therefore included success-imitation fine-tuning, instruction rewriting and execution calibration. Later generations retained this execution profile while changing action-head weights and instructions. In replication, the learned policies all received ``align to the target, then advance when centered''; the two Success-only conditions differed only in execution profile. IBVS does not consume a language instruction.

\section*{Supplementary Note 6: Benchtop experimental protocol}\label{si-note:6}

Benchtop insertion was tested with ER2 and a white silicone phantom containing a circular planar aperture 10~mm in diameter. At the protocol start pose, the distal tip was 5~cm from the aperture. The lateral offset was randomised between attempts while ensuring that the opening remained visible but was not centred in the image. Before each attempt, benchtop operator~1 disarmed the robot, returned the endoscope to this start pose, verified the 400$\times$400 RGB image stream and zero motor command, loaded the designated checkpoint and execution profile, and explicitly authorised motion. During an autonomous attempt, the policy alone generated motor commands; the operator could only terminate the trial or activate the emergency stop. The robot was disarmed and reset before the next attempt.

Under the S/F/E criterion, S required the endoscope tip to pass through the aperture plane under autonomous control. F comprised an attempt terminated without passage and without emergency-stop activation. E was recorded when the operator halted motion for an unsafe trajectory, abrupt advance or imminent equipment contact. Both F and E were unsuccessful outcomes. Each attempt received an operator-assigned terminal label at successful passage, declared failure or emergency stop. The archived review implementation sets \texttt{control.max\_trial\_seconds} to 180~s by default. Once elapsed time from arming exceeds this configurable threshold, the control loop zeros the motors and suppresses subsequent autonomous commands; it does not automatically assign F or E, close the trial or stop video recording. Every reported arm contained 21 attempts except ZS-A, which was halted after 16 consecutive emergency stops for safety. The fixed references included image-based visual servoing (IBVS). Supplementary Table~\ref{si-tab:phantomfull} gives every reported arm; trial-duration phenotypes are reported descriptively in Supplementary Table~\ref{si-tab:durationphenotypes}.

The three deployment-shift arms reused the Gen4 checkpoint without adaptation. For reduced illumination, the endoscope's integrated distal light source was switched off while ambient room lighting was held unchanged. The phantom-configuration shift retained the white silicone material and background but used a smaller 7-mm aperture adjacent to a protruding tumour-like distractor. The different-instrument arm replaced the 6.0-mm ER2 endoscope with the 3.9-mm ER3 endoscope while retaining the task definition.

\section*{Supplementary Note 7: Human--agent decision chronology}\label{si-note:7}

Supplementary Table~\ref{si-tab:provenance} reconstructs the recorded order of human decisions, coding-agent work and physical evidence. Validation-driven rejection of v1 preceded selection of continuous action chunks and temporal ensembling. Investigators then supplied the align-then-advance decomposition, and the agent tested source-task relations and prepared ZS-A, a dual-policy router and an instructed VLA candidate. The VLA manifest was fixed before ZS-A physical testing.

The update producing Gen2 used successful trajectories only. Failure-derived avoidance targets first entered the update producing Gen3 and were retained in the update producing Gen4. These changes are distinguished from instruction rewriting and execution calibration in the generation recipes.

\section*{Supplementary Note 8: Ex-vivo protocol and outcome assessment}\label{si-note:8}

The evaluation workflow, six tissue specimens and a successful Gen4 sequence are shown in Supplementary Fig.~\ref{si-fig:exvivoqual}. Supplementary Fig.~\ref{si-fig:exvivofail} shows unsuccessful Gen4 and Gen1 sequences. Each sequence is a qualitative illustration from an existing trial.

Six discarded porcine tracheae obtained after slaughter from a meat vendor at stall M36, Tai Po Hui Market, Tai Po, Hong Kong, were evaluated in two separate protocols. Specimens, rather than repeated insertions, are the independent biological units. The initial block cohort used specimens 1--3 and 84 attempts; the interleaved replication cohort used specimens 4--6 and 147 attempts. Tables retain the specimen-level cells, and every pooled Fisher comparison is explicitly exploratory at the level of technical attempts.

\subsection*{Initial block cohort}

The initial cohort was conducted during one 3-h-15-min session. Within a specimen, one policy completed a block of seven attempts before the next policy was loaded. For specimens 1 and 2, the block order was Gen4, Success-only, IBVS and Gen1; for specimen 3, it was Gen4, IBVS, Success-only and Gen1. The logged intervals were 04:00--06:08 for specimen 1, 06:17--06:47 for specimen 2 and 06:52--07:15 for specimen 3. Four policy arms were evaluated for seven attempts on each specimen. Thus, biological $n=3$ for this cohort, and the seven attempts in each policy--specimen cell are technical repetitions. All 84 protocol attempts entered the reported analysis. Two pre-experiment robot-movement checks, both ending by emergency stop, were outside the experimental protocol and did not enter any policy denominator; in particular, Gen4 remained 21/21. Initial ex-vivo operator~1's live record contained 31 ordinary failures and three emergency stops among the 34 unsuccessful protocol trials. All three emergency stops were analysed as failures.

For successful trials, navigation time was defined as the interval from navigation at one end of the ex-vivo segment to successful arrival at the other end. It is not the raw video-file duration. Supplementary Table~\ref{si-tab:exvivofull} reports the median of the available exact-path annotations for each success, then the within-cell median and interquartile range (IQR).

\subsection*{Interleaved replication cohort}

The independent replication cohort used three new tracheae (specimens 4--6). Seven conditions were evaluated: IBVS, Gen1, Gen4, the final Success-only checkpoint under its original execution configuration, the same Success-only checkpoint under the Gen4 execution configuration, experienced teleoperation and novice teleoperation. The intended schedule interleaved one attempt from every condition within each of seven rounds per specimen. The retained chronology reproduced the intended schedule except for a local four-position permutation in specimen 4; the complete schedule and trial manifest are supplied. Every condition still contributed seven attempts per specimen, giving 49 attempts per specimen and 147 protocol attempts.

The execution-configuration control held the final Success-only checkpoint fixed and changed only its runtime profile. Gen1 and original-configuration Success-only used the original profile; configured Success-only and Gen4 used the Gen4 profile. All learned policies received the same task instruction in replication. The exact profiles and common instruction are specified once in Supplementary Note~5. The configuration-control contrast therefore changes runtime settings for one checkpoint, whereas configured Success-only versus Gen4 changes the checkpoint under a common runtime profile.

\subsection*{Outcome assessment and human feedback}

Personnel are identified by experimental role rather than by name. Benchtop operator~1 conducted the silicone-phantom trials and supplied their live S/F/E outcomes; initial ex-vivo operator~1 conducted and recorded the first tissue cohort; and replication outcome recorder~1 entered the live outcomes for the second cohort. Experienced operator~1 had more than 100~h of prior robotic-endoscope control experience. Novice operator~1 had received only a system demonstration and setup introduction before the experiment. Video evaluator~1, video evaluator~2 and external evaluator~1 independently evaluated the replication videos. Video evaluator~1 also evaluated the 315 benchtop videos, video evaluator~2 evaluated the initial ex-vivo cohort, and external evaluator~1 evaluated all 546 scored videos.

The initial review used identity-free shuffled videos without policy or trial-folder metadata. Replication videos were judged independently using the same success criterion. Because emergency-stop activation was not observable from the endoscopic video alone, video-level E labels were analysed as failures and safety-event counts were taken from the live record.

For successful ex-vivo trials, each available video reviewer annotated the time from the start of navigation at one end of the tissue segment until successful arrival at the other end. The trial-level time was the median of the available exact-path annotations. Free text was optional; the two text-tagged records contributed no training sample. Terminal S/F/E outcomes supplied physical-outcome supervision. The coding agent can access human feedback on failure cases.

\subsection*{Ethics and tissue sourcing}

No live animals were used or killed for this study. The ex-vivo porcine stomach used for source tracking and the porcine tracheae used for transfer were discarded post-slaughter tissues obtained from the same vendor at stall M36, Tai Po Hui Market, Tai Po, Hong Kong. The study involved no human tissue, patients or clinical decisions. The two teleoperation conditions were descriptive system-reference arms, with one operator in each experience category.

\section*{Supplementary Note 9: Statistical analysis and rating reliability}\label{si-note:9}

\subsection*{Statistical analysis}

Benchtop binary proportions use two-sided Wilson 95\% CIs~\citep{si-wilson1927probable}. Two-arm binary comparisons use two-sided Fisher exact tests~\citep{si-fisher1922interpretation}. Principal benchtop contrasts were Gen4 versus Gen1, IBVS and final Success-only. Shift and trial-duration comparisons were descriptive (Supplementary Table~16). For ex-vivo transfer, the biological sample size was six specimens across two distinct protocols; the cohorts were analysed separately. Within the replication cohort, pooled Fisher exact comparisons treated repeated attempts as technical outcomes and are reported only as exploratory attempt-level analyses. Binary rating agreement used percentage agreement, pairwise Cohen's $\kappa$~\citep{si-cohen1960coefficient} and three-rater Fleiss' $\kappa$~\citep{si-fleiss1971measuring}. Success-time reliability used a two-way random-effects, absolute-agreement ICC for individual ratings, ICC(A,1), and for the mean of three ratings, ICC(A,$k$)~\citep{si-mcgraw1996forming}. Cohort-specific 95\% CIs were estimated from 10,000 trial-level bootstrap resamples using a fixed random seed for each cohort. Each teleoperation reference contains one operator, so no population-level inference about operator experience was performed.

Offline uncertainty used 20,000 paired bootstrap resamples of the 10 held-out episodes per task~\citep{si-efron1979bootstrap}. The paired difference was computed by first calculating the MAE difference within each episode and then averaging the 10 episode-level differences with equal weight; it therefore need not equal the difference between the two frame-pooled MAEs. Frames were never treated as independent replicates. Analysis used Python 3.10, NumPy 1.26 and SciPy 1.11. No a priori power calculation was performed; physical trial budgets reflected the experimental protocol, with ZS-A stopped early for safety.

\subsection*{Rating agreement and timing reliability}

Across the 315 video-scored benchtop trials (the 16 ZS-A safety-stop trials were not video scored) and both ex-vivo cohorts, the relevant live outcome and external evaluator~1 agreed on all 546 scored videos (Cohen's $\kappa=1.00$). The three rating sources selected per trial agreed on 546/546 exact-path matches (Fleiss' $\kappa=1.00$), including all 84 initial-cohort ex-vivo trials. In the replication cohort, every path matched and the live outcome plus all three video evaluations agreed on 147/147 binary outcomes. Binary outcome agreement is reported in Supplementary Table~17.

In the initial cohort, 48 successful trials had complete three-rater timing annotations and the absolute-agreement intraclass correlation coefficient (ICC) for an individual rating, ICC(A,1), was 0.994 (bootstrap 95\% CI, 0.978--0.998). In the replication cohort, all 76 successful trials had complete three-rater timings: ICC(A,1) was 0.956 (0.873--0.984), and the coefficient for the mean of three ratings, ICC(A,$k$), was 0.985 (0.954--0.995). Timing-reliability estimates and flagged records are reported in Supplementary Table~18.

\clearpage\section*{Supplementary Figures}

\clearpage

\begin{figure}[!htbp]
\centering
\includegraphics[width=\linewidth,height=0.75\textheight,keepaspectratio]{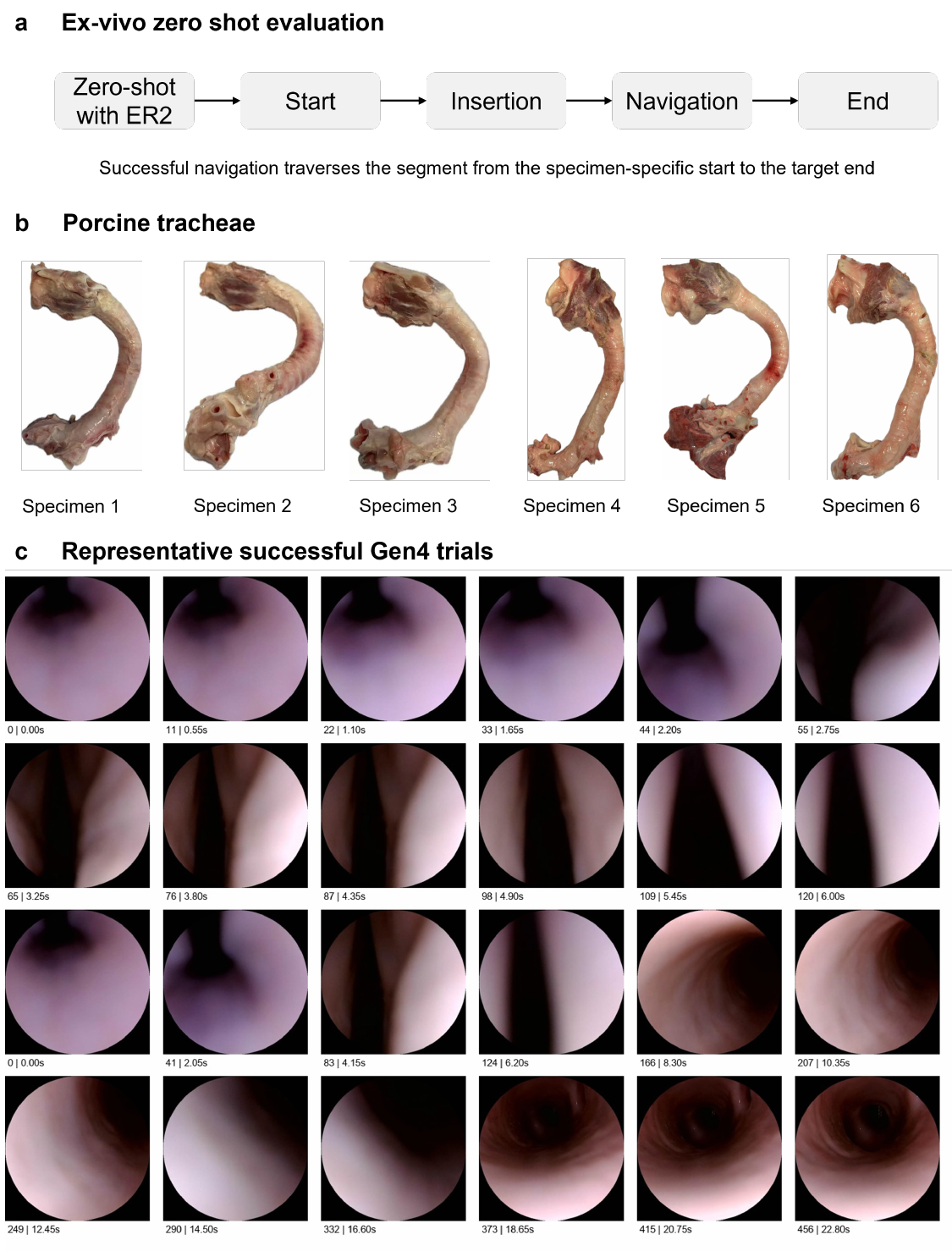}
\caption{\textbf{Ex-vivo evaluation and a successful Gen4 trial.} \textbf{a}, ER2 evaluation workflow. Here, zero-shot denotes transfer of the fixed policy to ex-vivo tissue without adaptation. \textbf{b}, Six independent porcine tracheae: specimens 1--3 from the initial block cohort and specimens 4--6 from the interleaved replication cohort; no calibrated scale is implied. \textbf{c}, One successful Gen4 trial on specimen~4 from the interleaved replication cohort. The first two rows sample the first 6.00~s; the next two sample the full recording through 22.80~s. These are two temporal samplings of the same trial, with repeated frames and a reset time axis. Labels give zero-based frame indices and elapsed video times at 20~frames~s$^{-1}$, not scored traversal times. This qualitative example adds no trials or biological replicates.}
\label{si-fig:exvivoqual}
\end{figure}
\clearpage

\begin{figure}[!htbp]
\centering
\includegraphics[width=\linewidth,height=0.77\textheight,keepaspectratio]{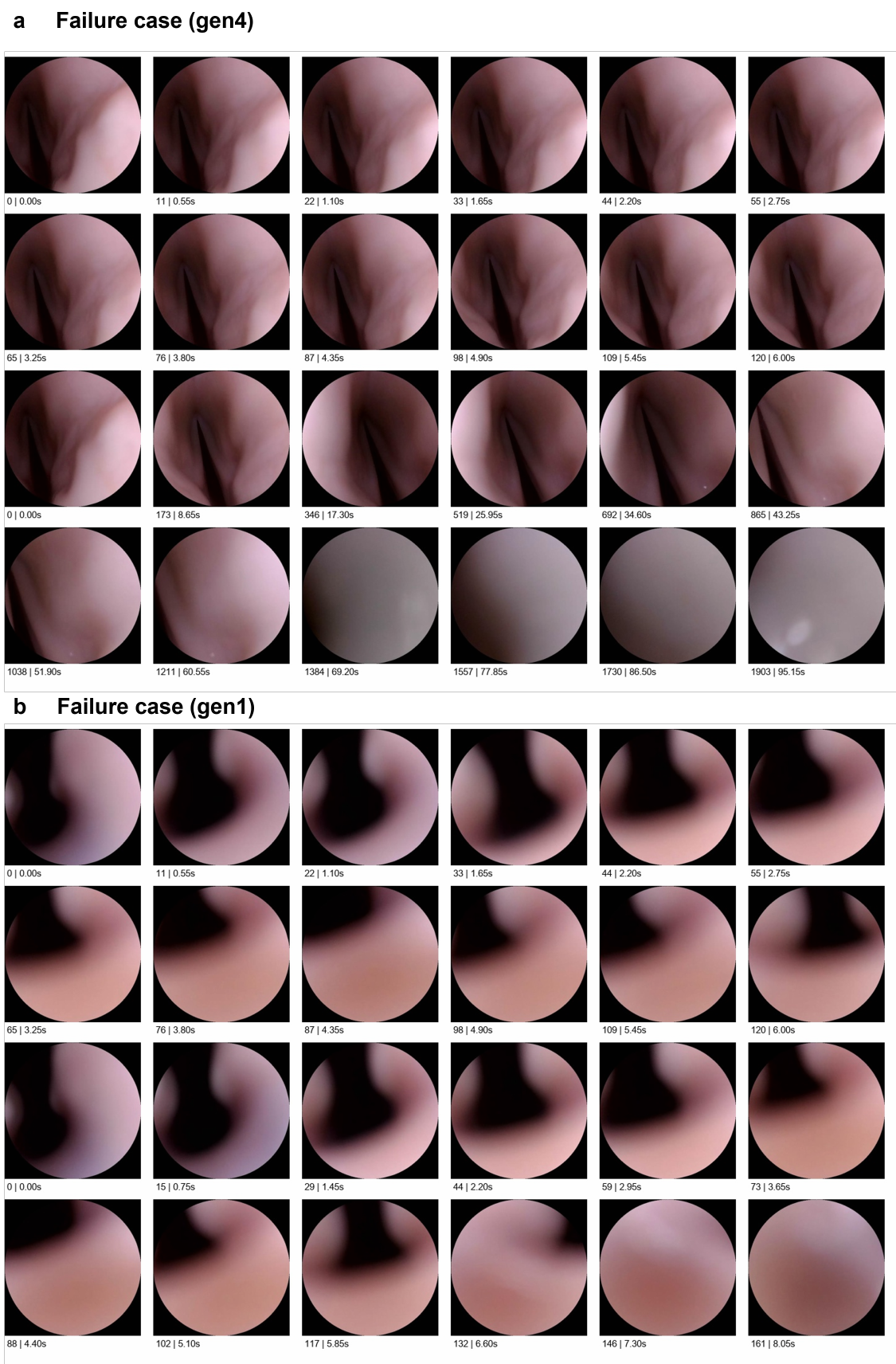}
\caption{\textbf{Representative unsuccessful ex-vivo trials.} \textbf{a}, Gen4 on specimen~6 from the interleaved replication cohort. \textbf{b}, Gen1 on specimen~2 from the initial block cohort. In each panel, the first two rows sample the first 6.00~s, and the next two sample the full recording, through 95.15~s in \textbf{a} and 8.05~s in \textbf{b}. Each panel shows one trial with two temporal samplings, repeated frames and a reset time axis. Labels give zero-based frame indices and elapsed video times at 20~frames~s$^{-1}$. Both trials have unsuccessful canonical outcomes; the terminal views do not establish successful lumen alignment. The images alone do not establish failure mechanism or emergency-stop activation. These qualitative examples add no trials or biological replicates.}
\label{si-fig:exvivofail}
\end{figure}
\clearpage

\section*{Supplementary Tables}

\begin{table}[!htbp]
\centering
\caption{Implemented evidence-to-policy lifecycle.}
\label{si-tab:lifecycle}
\small
\begin{tabularx}{\linewidth}{C{0.07\linewidth} C{0.16\linewidth} Y Y}
\toprule
Stage & Scientific role & Evidence and decision & Frozen output \\
\midrule
1 & Audit source skills & Verify image--action correspondence, episode splits, ranges and task statistics & Locked splits, normalisation and audit report \\
2 & Test policy hypotheses & Reject v1 on validation; replace discrete residual actions with continuous chunks and temporal ensembling (TE) & Selected source-policy checkpoint and validation report \\
3 & Operationalise a task hypothesis & Human specifies align-then-advance; agent quantifies source-task support, rejects unsupported navigation-centroid steering and implements three candidates & Cross-task analysis; zero-shot composition A (ZS-A), dual-policy router and vision--language--action (VLA) manifests \\
4 & Falsify physically & Sixteen consecutive ZS-A emergency stops reject discontinuous rule composition & Sealed outcome set and transition to reactive policy \\
5 & Develop reactive insertion & Agent develops its pre-frozen VLA direction after physical falsification and a human requirement for end-to-end stop-capable behaviour & Gen1 checkpoint and manifest \\
6 & Evolve from evidence & Gen1 $\rightarrow$ Gen2 clones success only; failure relabelling first enters the Gen2 $\rightarrow$ Gen3 update & Gen2--Gen4 outcome records, recipes and research ledger \\
7 & Test transfer & Freeze Gen4; evaluate shifts and discarded porcine tracheae without adaptation & Shifted-condition and ex-vivo trial records \\
\bottomrule
\end{tabularx}
\end{table}
\FloatBarrier

\begin{table}[!htbp]
\centering
\caption{Control flow and safety authority. The coding agent acts between trials and has no camera or motor interface while the robot is armed. S, F and E denote success, ordinary failure and emergency stop.}
\label{si-tab:controlpath}
\small
\begin{tabularx}{\linewidth}{p{0.19\linewidth} Y Y}
\toprule
Component & Input and operation & Output and authority \\
\midrule
Human investigators & Define the task and constraints; reset and authorise each trial & Retain deployment and emergency-stop authority; assign terminal S/F/E outcomes \\
Coding agent & Audit data and results; implement and revise candidates between trials & Prepare a future checkpoint, manifest, instruction and execution limits for human authorisation \\
Frozen candidate & One authorised checkpoint, manifest, instruction and limit configuration & Remains fixed throughout the armed trial \\
Visual processing & Full endoscopic frames pass through preprocessing and asynchronous inference & Semantic action predictions enter an action queue \\
Execution runtime & The 20-Hz loop selects valid queued actions & Deterministic magnitude, watchdog and slew limits bound the command \\
Platform adapter & Convert semantic commands to the selected platform's motor mapping & Send platform-specific packets to serial firmware \\
Physical execution & Firmware actuates the motors; the camera closes the visual loop & Human emergency stop overrides motor output \\
Trial evidence & Preserve outcomes and recorded observations after each trial & Return evidence to the coding workspace for subsequent analysis and revision \\
\bottomrule
\end{tabularx}
\end{table}
\FloatBarrier

\begin{table}[!htbp]
\centering
\caption{Instrument, experiment and controller mapping.}
\label{si-tab:platforms}
\small
\begin{tabularx}{\linewidth}{C{0.08\linewidth} C{0.23\linewidth} Y C{0.17\linewidth}}
\toprule
Platform & Instrument & Dataset or experiment role & Lower controller \\
\midrule
ER1 & Olympus GIF-Q260J medical gastroscope, 9.9 mm & Tracking source dataset in discarded ex-vivo pig stomach & Four-motor firmware \\
ER2 & Ailishi S50 industrial endoscope, 6.0 mm & Navigation source dataset; development and evolution aperture; low illumination; phantom-configuration shift; both ex-vivo cohorts & Six-channel firmware (four motors are used); M1--M4 active, M5/M6 zero \\
ER3 & Ailishi S50 industrial endoscope, 3.9 mm & Different-instrument transfer arm & Same six-channel firmware (four motors are used) and active-channel pattern as ER2 \\
\bottomrule
\end{tabularx}
\end{table}
\FloatBarrier

\begin{table}[!htbp]
\centering
\caption{Relationship among collection data, learned policy and hardware code.}
\label{si-tab:codepath}
\small
\begin{tabularx}{\linewidth}{C{0.18\linewidth} Y Y}
\toprule
Layer & Input and output & Reproducible role in the archive \\
\midrule
Platform collection & Camera, timestamp and executed device command $\rightarrow$ recorded episode & Three device-facing collection paths preserve the raw embodiment and acquisition context \\
Dataset conversion & Recorded episode $\rightarrow$ aligned image--semantic-action records & Validates image/action correspondence, constructs episode-disjoint splits and converts commands to $[m_1,m_3,m_4]$ \\
Training & Canonical records and task normalisation $\rightarrow$ frozen checkpoint and manifest & Learns the action controller without embedding a raw ER1/ER2/ER3 serial packet \\
Inference runtime & Current image, instruction and frozen manifest $\rightarrow$ bounded semantic action & Runs asynchronous model inference and a deterministic 20-Hz watchdog/safety boundary \\
Hardware adapter & Semantic action plus platform profile $\rightarrow$ device motor packet & Applies channel expansion, sign, scale, gain, bias and inactive-channel rules \\
Lower firmware & Device motor packet $\rightarrow$ motor actuation & Separate four-motor medical and six-channel industrial Keil projects \\
\bottomrule
\end{tabularx}
\end{table}
\FloatBarrier

\begin{table}[!htbp]
\centering
\caption{Source-dataset audit and experimental units.}
\label{si-tab:datasets}
\small
\begin{tabularx}{\linewidth}{YrrrrY}
\toprule
Task & Episodes & Action rows & Split & Test $n$ & Collection system and environment \\
\midrule
Tracking & 100 & 12,710 & 80/10/10 & 10 episodes & ER1, Olympus GIF-Q260J; ex-vivo pig stomach \\
Navigation & 100 & 16,534 & 80/10/10 & 10 episodes & ER2, Ailishi S50 6.0 mm; colon phantom \\
\bottomrule
\end{tabularx}
\end{table}
\FloatBarrier

\begin{table}[!htbp]
\centering
\caption{Validation-based model selection and post-selection held-out performance. Values are frame-pooled mean absolute error (MAE) in raw motor units; lower is better. Each task contains 100 episodes split into 80 training, 10 validation and 10 test episodes. Validation determined candidate selection; held-out scores were evaluated afterwards. TE denotes causal temporal ensembling of the selected v2 checkpoint and is a decoding procedure, not a separately trained model.}
\label{si-tab:offlineselection}
\small
\begin{tabular}{lrrrr}
\toprule
& \multicolumn{2}{c}{Tracking} & \multicolumn{2}{c}{Navigation} \\
\cmidrule(lr){2-3}\cmidrule(lr){4-5}
Policy & Validation & Held-out & Validation & Held-out \\
\midrule
CNN behaviour cloning & 2.619 & 2.316 & 19.463 & 23.580 \\
Agent v1 & 2.654 & 2.065 & 19.811 & 23.608 \\
Agent v2 & 2.615 & 2.255 & 19.091 & 22.032 \\
Agent v2+TE & 2.426 & 2.112 & 18.791 & 21.683 \\
\bottomrule
\end{tabular}
\par\medskip
\begin{minipage}{\linewidth}
\small
The v1 representation was rejected because validation MAE exceeded the CNN reference on both tasks. Direct continuous prediction and TE reduced validation error. The complete held-out image-conditioned comparison, including additional references, RMSE and sign accuracy, is reported in Supplementary Table~\ref{si-tab:offline}; three-run stability is reported in Supplementary Table~\ref{si-tab:seeds}. Paired-bootstrap uncertainty uses equally weighted episode-level differences, rather than frame-pooled scores. The teacher-forced copy-previous diagnostic is excluded from image-conditioned comparisons.
\end{minipage}
\end{table}
\FloatBarrier

\begin{table}[!htbp]
\centering
\caption{Complete held-out image-conditioned offline results. Lower MAE and RMSE are better; higher sign accuracy is better. The copy-previous diagnostic is excluded because it is teacher forced.}
\label{si-tab:offline}
\small
\begin{tabular}{llrrr}
\toprule
Task & Policy & MAE & RMSE & Sign accuracy \\
\midrule
Tracking & Classical visual servoing & 2.846 & 4.006 & 0.778 \\
 & Diffusion Policy & 2.340 & 3.891 & 0.780 \\
 & Action-chunking transformer & 2.417 & 3.830 & 0.781 \\
 & CNN behaviour cloning & 2.316 & 3.697 & 0.817 \\
 & Image-only VLA & 2.622 & 4.223 & 0.611 \\
 & v1 & 2.065 & 3.884 & 0.741 \\
 & v2 & 2.255 & 3.538 & 0.808 \\
 & v2+TE & 2.112 & 3.319 & 0.820 \\
\midrule
Navigation & Diffusion Policy & 24.422 & 42.955 & 0.844 \\
 & Action-chunking transformer & 23.751 & 42.127 & 0.835 \\
 & CNN behaviour cloning & 23.580 & 41.820 & 0.881 \\
 & Image-only VLA & 26.038 & 44.471 & 0.782 \\
 & v1 & 23.608 & 45.402 & 0.826 \\
 & v2 & 22.032 & 38.927 & 0.886 \\
 & v2+TE & 21.683 & 38.110 & 0.910 \\
\bottomrule
\end{tabular}
\end{table}
\FloatBarrier

\begin{table}[!htbp]
\centering
\caption{Selected-policy stability across three training runs. Values are held-out test MAE; s.d. denotes standard deviation.}
\label{si-tab:seeds}
\small
\begin{tabular}{lrrrrr}
\toprule
Task & Original & Seed 1 & Seed 2 & Mean $\pm$ s.d. & CNN reference \\
\midrule
Tracking & 2.112 & 2.076 & 1.995 & $2.061\pm0.060$ & 2.316 \\
Navigation & 21.683 & 21.257 & 21.615 & $21.518\pm0.229$ & 23.580 \\
\bottomrule
\end{tabular}
\end{table}
\FloatBarrier

\begin{table}[!htbp]
\centering
\caption{Source-task evidence and its disposition.}
\label{si-tab:cross}
\small
\begin{tabularx}{\linewidth}{YrrY}
\toprule
Claim or candidate & Sample size & Evidence & Decision \\
\midrule
Tracking centre error predicts bending & 10,284 frames & $R^2=0.340$ ($m_3$), 0.370 ($m_4$) & Retain as alignment primitive \\
Navigation dark-centroid error predicts bending & 13,810 frames & $R^2=-0.034$ ($m_3$), 0.004 ($m_4$) & Reject \\
Low bending predicts advancement & 13,810 frames & Mean $m_1=16.746$ in lowest quartile, $\leq0.560$ otherwise & Retain as conditional advance principle \\
ZS-A composes both rules safely & 16 physical attempts & 0 success; 16 emergency stops & Physical falsification rejects rule composition \\
\bottomrule
\end{tabularx}
\end{table}
\FloatBarrier

\begin{table}[!htbp]
\centering
\caption{Generation-specific development evidence and update recipes. Source-batch outcomes are recorded in the supplied per-trial outcome correspondence; the last column gives the final video-reviewed performance of the resulting policy. Comparisons between lineages are descriptive because their evidence pools and starting checkpoints differed.}
\label{si-tab:trainingrules}
\small
\begin{tabularx}{\linewidth}{C{0.16\linewidth} Y Y C{0.14\linewidth}}
\toprule
Update & Source-batch outcomes & Training use & Successor outcome \\
\midrule
Gen1 $\rightarrow$ Gen2 & 2 success; 19 failure/emergency stop & Success only; 175 cloning samples; 300 steps; loss 0.4662 $\rightarrow$ 0.2851 & 8/21 \\
Gen2 $\rightarrow$ Gen3 & 8 success; 13 failure/emergency stop & Failure-derived: 68 cloning + 144 mirrored avoidance samples from 12 contributing records; nine records supplied no sample; loss 0.2243 $\rightarrow$ 0.0918 & 14/21 \\
Gen3 $\rightarrow$ Gen4 & 14 success; 7 failure & Failure-derived: 240 cloning + 160 avoidance samples; 300 steps; loss 0.0914 $\rightarrow$ 0.0643 & 17/21 \\
Success-only control & 1/21 $\rightarrow$ 3/21 $\rightarrow$ 3/21 $\rightarrow$ 3/21 & Each update used successes from its own preceding batch and discarded unsuccessful outcomes & 3/21 final \\
\bottomrule
\end{tabularx}
\end{table}
\FloatBarrier

\begin{table}[!htbp]
\centering
\caption{Complete benchtop outcomes with exact sample sizes and Wilson 95\% confidence intervals (CIs). Emergency-stop counts come from benchtop operator~1's live record; endoscopic video alone does not establish emergency-stop activation.}
\label{si-tab:phantomfull}
\small
\begin{tabular}{llrrr}
\toprule
Group & Arm & Success / $n$ & Success, \% (95\% CI) & Emergency stop \\
\midrule
Physical falsification & ZS-A & 0/16 & 0.0 (0.0--19.4) & 16 \\
\midrule
Fixed reference & CNN behaviour cloning & 0/21 & 0.0 (0.0--15.5) & 0 \\
 & Action-chunking transformer & 1/21 & 4.8 (0.8--22.7) & 1 \\
 & Diffusion Policy & 2/21 & 9.5 (2.7--28.9) & 0 \\
 & IBVS & 3/21 & 14.3 (5.0--34.6) & 0 \\
\midrule
Success-only & Iteration 1 & 1/21 & 4.8 (0.8--22.7) & 0 \\
 & Iteration 2 & 3/21 & 14.3 (5.0--34.6) & 0 \\
 & Iteration 3 & 3/21 & 14.3 (5.0--34.6) & 0 \\
 & Iteration 4 & 3/21 & 14.3 (5.0--34.6) & 0 \\
\midrule
Generation sequence & Gen1 & 2/21 & 9.5 (2.7--28.9) & 5 \\
 & Gen2 & 8/21 & 38.1 (20.8--59.1) & 0 \\
 & Gen3 & 14/21 & 66.7 (45.4--82.8) & 0 \\
 & Gen4 & 17/21 & 81.0 (60.0--92.3) & 0 \\
\midrule
Frozen Gen4 shift & Low illumination & 13/21 & 61.9 (40.9--79.2) & 0 \\
 & Phantom-configuration shift & 13/21 & 61.9 (40.9--79.2) & 0 \\
 & Different instrument & 15/21 & 71.4 (50.0--86.2) & 0 \\
\bottomrule
\end{tabular}
\end{table}
\FloatBarrier

\begin{table}[!htbp]
\centering
\caption{Two-sided Fisher exact tests for final phantom contrasts and descriptive generation comparisons. Generation comparisons quantify observed proportions; causal interpretation is limited because the Success-only and generation lineages used different evidence pools and starting checkpoints.}
\label{si-tab:phantomstats}
\small
\begin{tabularx}{\linewidth}{YrrY}
\toprule
Comparison & Odds ratio & $P$ & Role \\
\midrule
Gen4 versus Gen1 & 40.38 & $5.75\times10^{-6}$ & Overall generation contrast \\
Gen4 versus IBVS & 25.50 & $3.21\times10^{-5}$ & Best fixed classical reference \\
Gen4 versus final Success-only iteration & 25.50 & $3.21\times10^{-5}$ & Failure-evidence mechanism control \\
Gen2 versus Gen1 & 5.85 & 0.0670 & Generation comparison \\
Gen3 versus Gen2 & 3.25 & 0.121 & Generation comparison \\
Gen4 versus Gen3 & 2.13 & 0.484 & Generation comparison \\
Success-only iteration 4 versus iteration 1 & 3.33 & 0.606 & Repeated success-only updating \\
Gen4 versus low illumination & 2.62 & 0.306 & Descriptive deployment-shift comparison \\
Gen4 versus phantom-configuration shift & 2.62 & 0.306 & Descriptive deployment-shift comparison \\
Gen4 versus different instrument & 1.70 & 0.719 & Descriptive deployment-shift comparison \\
\bottomrule
\end{tabularx}
\end{table}
\FloatBarrier

\begin{table}[!htbp]
\centering
\small
\setlength{\tabcolsep}{5pt}
\caption{Human--agent division of scientific decisions. Rows follow the recorded order of research decisions. Investigators retain deployment and emergency-stop authority. ZS-B denotes the dual-policy router.}\label{si-tab:provenance}
\begin{tabularx}{\linewidth}{Y Y}
\toprule
Human investigators & Coding agent \\
\midrule
Define research goal, align-then-advance task and bounded deployment authority. & Plan dataset audit, validation-based revision and physical evaluation. \\[4pt]
Provide existing tracking and navigation datasets and experimental constraints. & Audit image-action records, lock episode splits and implement comparison policies. \\[4pt]
Require validation evidence to guide policy improvement. & Reject v1 on validation. Implement continuous action chunks and causal temporal ensembling. \\[4pt]
Specify insertion as tracking plus navigation. Require freezing before insertion outcomes. & Quantify cross-task support. Reject centroid steering. Freeze ZS-A, ZS-B and instructed VLA. \\[4pt]
Authorise ZS-A tests and use emergency stop in all 16 attempts. & Analyse discontinuous advancement and absence of a reactive stop response. \\[4pt]
Prioritise end-to-end VLA development and specify stop-on-contact behaviour. & Develop the previously frozen VLA direction into Gen1 using source-derived supervision. \\[4pt]
Authorise each fixed candidate, reset trials and assign S / F / E terminal outcomes. & Gen1 to Gen2: success imitation. Gen2 to Gen3 and Gen3 to Gen4: failure-derived supervision. \\[4pt]
Run four autonomous policies on three independent tracheae, seven attempts per cell. & Keep Gen4 fixed. Summarise specimen-level results and repeated technical attempts. \\[4pt]
Run the interleaved seven-condition protocol on three new tracheae. & Evaluate fixed policies and configuration controls, including teleoperation references. \\[4pt]
\bottomrule
\end{tabularx}
\end{table}
\FloatBarrier

\begin{table}[!htbp]
\centering
\caption{Initial-cohort ex-vivo outcomes and successful navigation times by policy and specimen. The specimen is the biological unit; attempts are repeated technical measurements. F and E are distinct live-operator events but both are binary failures. Navigation time is conditional on success and shown as median [IQR].}
\label{si-tab:exvivofull}
\small
\begin{tabular}{llrrrrl}
\toprule
Policy & Specimen & $n$ & S & F & E & Success navigation time, s \\
\midrule
IBVS & 1 & 7 & 3 & 3 & 1 & 6.2 [4.9--14.6] \\
 & 2 & 7 & 3 & 3 & 1 & 2.0 [1.9--2.1] \\
 & 3 & 7 & 3 & 4 & 0 & 2.5 [1.9--3.2] \\
\midrule
Gen1 & 1 & 7 & 2 & 5 & 0 & 9.3 [8.9--9.6] \\
 & 2 & 7 & 4 & 3 & 0 & 3.5 [2.4--4.5] \\
 & 3 & 7 & 3 & 3 & 1 & 6.7 [4.7--6.7] \\
\midrule
Success-only & 1 & 7 & 3 & 4 & 0 & 18.3 [14.7--20.4] \\
 & 2 & 7 & 4 & 3 & 0 & 3.1 [2.9--3.8] \\
 & 3 & 7 & 4 & 3 & 0 & 4.6 [3.2--7.0] \\
\midrule
Gen4 & 1 & 7 & 7 & 0 & 0 & 11.9 [10.3--17.3] \\
 & 2 & 7 & 7 & 0 & 0 & 10.3 [8.4--12.7] \\
 & 3 & 7 & 7 & 0 & 0 & 6.6 [3.8--8.0] \\
\bottomrule
\end{tabular}
\end{table}
\FloatBarrier

\begin{table}[!htbp]
\centering
\caption{Cohort-specific pooled descriptive outcomes and exploratory attempt-level comparisons. Each cohort contains three independent specimens and seven technical repetitions per condition--specimen cell. Cohorts are analysed separately. Wilson 95\% CIs are descriptive; two-sided Fisher tests compare autonomous controls with Gen4 within the same cohort and do not provide specimen-level or causal inference. The initial Gen4 row includes its 21 protocol attempts only. Teleoperation rows are single-operator references and were not tested for an experience effect.}
\label{si-tab:exvivosensitivity}
\small
\begin{tabularx}{\linewidth}{YrrrY}
\toprule
Condition & Success / $n$ & Success, \% (95\% CI) & Exploratory $P$ vs Gen4 & Scope \\
\midrule
\multicolumn{5}{l}{Initial block cohort} \\
\midrule
Gen4 & 21/21 & 100.0 (84.5--100.0) & -- & Frozen evolved policy \\
Success-only & 11/21 & 52.4 (32.4--71.7) & $4.79\times10^{-4}$ & Final success-only control \\
Gen1 & 9/21 & 42.9 (24.5--63.5) & $5.32\times10^{-5}$ & Initial reactive policy \\
IBVS & 9/21 & 42.9 (24.5--63.5) & $5.32\times10^{-5}$ & Classical reference \\
\midrule
\multicolumn{5}{l}{Interleaved replication cohort} \\
\midrule
Gen4 & 16/21 & 76.2 (54.9--89.4) & -- & Frozen evolved checkpoint \\
Success-only, Gen4 configuration & 7/21 & 33.3 (17.2--54.6) & 0.0122 & Checkpoint contrast under a common runtime profile \\
Success-only, original configuration & 4/21 & 19.0 (7.7--40.0) & $5.06\times10^{-4}$ & Original checkpoint/configuration system \\
Gen1 & 5/21 & 23.8 (10.6--45.1) & 0.00168 & Initial checkpoint under its original runtime profile \\
IBVS & 2/21 & 9.5 (2.7--28.9) & $2.49\times10^{-5}$ & Classical autonomous reference \\
Experienced teleoperation & 21/21 & 100.0 (84.5--100.0) & -- & Single experienced-operator system reference \\
Novice teleoperation & 21/21 & 100.0 (84.5--100.0) & -- & Single novice-operator system reference \\
\bottomrule
\end{tabularx}
\end{table}
\FloatBarrier

\begin{table}[!htbp]
\centering
\caption{Descriptive video-duration phenotypes for Success-only and generation arms. Duration includes the recorded trial video and is influenced by operator termination; it is not a time-to-event efficacy analysis. Median values are shown with IQR. Emergency-stop counts come only from benchtop operator~1's live record.}
\label{si-tab:durationphenotypes}
\small
\begin{tabular}{llrrrrr}
\toprule
Arm & Outcome & $n$ & Mean, s & Median [IQR], s & Range, s & Emergency stop \\
\midrule
Success-only 1 & F & 20 & 4.6 & 4.1 [3.6--5.0] & 3.3--10.2 & 0 \\
Success-only 2 & F & 18 & 3.5 & 3.2 [3.1--3.7] & 2.5--7.0 & 0 \\
Success-only 3 & F & 18 & 3.8 & 3.5 [3.2--4.2] & 3.0--5.6 & 0 \\
Success-only 4 & F & 18 & 4.8 & 4.2 [3.5--4.6] & 2.7--13.9 & 0 \\
\midrule
Gen1 & F+E & 19 & 12.4 & 5.3 [3.8--6.9] & 1.9--97.8 & 5 \\
Gen2 & F & 13 & 9.2 & 8.0 [6.9--9.9] & 3.4--21.2 & 0 \\
Gen3 & F & 7 & 29.2 & 15.1 [12.1--17.3] & 8.9--122.1 & 0 \\
Gen4 & F & 4 & 57.1 & 40.7 [21.6--76.2] & 8.0--138.9 & 0 \\
Gen4 & S & 17 & 50.7 & 49.4 [27.4--67.0] & 13.6--128.2 & 0 \\
\bottomrule
\end{tabular}
\end{table}
\FloatBarrier

\begin{table}[!htbp]
\centering
\caption{Binary outcome agreement after analysing every E outcome as F. The scored set comprises 315 phantom trials, 84 initial-cohort ex-vivo trials and 147 replication-cohort trials. The three-way analysis uses three rating sources per trial: live record, video evaluator~1 and external evaluator~1 for benchtop trials; live record, video evaluator~2 and external evaluator~1 for the initial cohort; and the three video evaluators for replication. Emergency-stop counts are not inferred from video and therefore come only from the relevant live record. Exact-path matching is used, with complete binary outcome ratings from the three selected sources for all scored trials.}
\label{si-tab:rateragreement}
\small
\begin{tabularx}{\linewidth}{YYrrY}
\toprule
Domain & Comparison & Exact-path $n$ & Agreement, \% & Reliability \\
\midrule
All scored trials & Relevant live record vs external evaluator 1 & 546 & 100.0 & Cohen's $\kappa=1.00$ \\
All scored trials & Three rating sources per trial & 546 & 100.0 & Fleiss' $\kappa=1.00$ \\
Phantom & Benchtop operator 1 vs video evaluator 1 & 315 & 100.0 & Cohen's $\kappa=1.00$ \\
Phantom & Benchtop operator 1 vs external evaluator 1 & 315 & 100.0 & Cohen's $\kappa=1.00$ \\
Phantom & All three rating sources & 315 & 100.0 & Fleiss' $\kappa=1.00$ \\
Initial ex vivo & Initial ex-vivo operator 1 vs video evaluator 2 & 84 & 100.0 & Cohen's $\kappa=1.00$ \\
Initial ex vivo & Initial ex-vivo operator 1 vs external evaluator 1 & 84 & 100.0 & Cohen's $\kappa=1.00$ \\
Initial ex vivo & All three rating sources & 84 & 100.0 & Fleiss' $\kappa=1.00$ \\
Replication ex vivo & Replication outcome recorder 1 vs each video evaluator & 147 each & 100.0 & Three pairwise $\kappa=1.00$ \\
Replication ex vivo & All three video raters & 147 & 100.0 & Fleiss' $\kappa=1.00$ \\
\bottomrule
\end{tabularx}
\end{table}
\FloatBarrier

\begin{table}[!htbp]
\centering
\caption{Reliability of successful ex-vivo navigation-time annotations. The intraclass correlation coefficient (ICC) is reported as ICC(A,1) for two-way random-effects absolute agreement of a single rating and as ICC(A,$k$) for reliability of the three-rating mean. ICC 95\% CIs use 10,000 trial-level bootstrap resamples with a fixed random seed for each cohort.}
\label{si-tab:timingreliability}
\small
\begin{tabularx}{\linewidth}{YYrrrY}
\toprule
Cohort & Analysis & $n$ & Estimate & 95\% CI & Interpretation \\
\midrule
Initial block & ICC(A,1) & 48 & 0.994 & 0.978--0.998 & Individual-rating agreement \\
Initial block & ICC(A,$k$) & 48 & 0.998 & 0.993--0.999 & Three-rating mean reliability \\
Interleaved replication & ICC(A,1) & 76 & 0.956 & 0.873--0.984 & Individual-rating agreement \\
Interleaved replication & ICC(A,$k$) & 76 & 0.985 & 0.954--0.995 & Three-rating mean reliability \\
\bottomrule
\end{tabularx}
\end{table}
\FloatBarrier

\begin{table}[!htbp]
\centering
\caption{Transfer outcomes across the two ex-vivo cohorts and six porcine tracheae. Specimens are independent biological units; each specimen cell contains seven technical repetitions, and Total aggregates the three cells within a cohort. Each cohort block lists only conditions evaluated in that protocol. Within replication, the two Success-only rows use the same checkpoint and differ only in the runtime execution profile: action scaling, gains and smoothing (Supplementary Note~5).}
\label{si-tab:exvivocombined}
\small
\begin{tabularx}{\linewidth}{Yrrrr}
\toprule
\multicolumn{5}{l}{Initial block cohort} \\
\midrule
Protocol condition & Specimen 1 & Specimen 2 & Specimen 3 & Total \\
\midrule
IBVS & 3/7 & 3/7 & 3/7 & 9/21 \\
Gen1 & 2/7 & 4/7 & 3/7 & 9/21 \\
Success-only, original configuration & 3/7 & 4/7 & 4/7 & 11/21 \\
Gen4 & 7/7 & 7/7 & 7/7 & 21/21 \\
\midrule
\multicolumn{5}{l}{Interleaved replication cohort} \\
\midrule
Protocol condition & Specimen 4 & Specimen 5 & Specimen 6 & Total \\
\midrule
IBVS & 2/7 & 0/7 & 0/7 & 2/21 \\
Gen1 & 2/7 & 2/7 & 1/7 & 5/21 \\
Success-only, original configuration & 2/7 & 2/7 & 0/7 & 4/21 \\
Success-only, Gen4 configuration & 2/7 & 2/7 & 3/7 & 7/21 \\
Gen4 & 6/7 & 5/7 & 5/7 & 16/21 \\
Experienced operator 1, teleoperation & 7/7 & 7/7 & 7/7 & 21/21 \\
Novice operator 1, teleoperation & 7/7 & 7/7 & 7/7 & 21/21 \\
\bottomrule
\end{tabularx}
\end{table}
\FloatBarrier

\clearpage\renewcommand{\refname}{Supplementary References}
\begingroup\small
\phantomsection\label{si-si:references}

\endgroup

\end{document}